%% file: paper.tex
\ifdefined\pdfminorversion
\fi

\documentclass[]{jingdong}

\input{common}

\title{Long-Horizon Audio-Visual Generation for Persistent Stories and Interactive Worlds}

\author{%
Nan Duan \quad Haoyang Huang \quad Weiyang Jin \quad Haoran Li \quad Yaowei Li \quad Yuming Li \\
Yijun Liu \quad Xin Lu \quad Xiaoxiao Ma \quad Yanwen Ma \quad Yaofeng Su \\
Yilang Sun \quad Haoyu Wang \quad Zeyue Xue \quad Songchun Zhang \quad Junhao Zhuang \\[0.6em]
{\normalfont\textit{Joy Future Academy, JD}}
}

\input{abstract}

\checkdata[Project Page]{\url{https://echo-team-joy-future-academy-jd.github.io/Echo-1.5-Page/}}
\checkdata[Code]{\url{https://github.com/jd-opensource/JoyAI-Echo}}

\begin{document}

\maketitle

\input{sections/1_intro}
\input{sections/2_Data}

\input{sections/3_echolong}
\input{sections/4_wm}

\input{sections/5_ar}
\input{sections/6_agent}
\input{sections/7_sr}
\input{sections/8_eval}

\input{conclusion}

\clearpage
\section{Authors}
\subsection{Core Contributors}
Nan Duan, Haoyang Huang, Weiyang Jin, Haoran Li, Yaowei Li, Yuming Li, Yijun Liu, Xin Lu, Xiaoxiao Ma, Yanwen Ma, Yaofeng Su, Yilang Sun, Haoyu Wang, Zeyue Xue, Songchun Zhang, Junhao Zhuang

\subsection{Contributors}
Bi Cheng, Jiayi Deng, Zuopeng Dong, Libing Fang, Tong He, Xintong Liu, Ruofan Lv, Jiaqi Shi, Zihan Su, Jiaqi Wang, Meihui Wang, Pan Wang, Haoyu Wu, Shiyi Zhang, Xuying Zhang, Mingchen Zhong

\clearpage
\bibliographystyle{plainnat}
\bibliography{cite}

\end{document}

%% file: common.tex
\usepackage{iftex}
\ifPDFTeX
  \usepackage[T1]{fontenc}
  \usepackage[utf8]{inputenc}
\fi

\usepackage{amsmath}
\usepackage{amssymb}
\usepackage{amsfonts}
\usepackage{nicefrac}
\usepackage{enumitem}
\usepackage{tabularx}

\AtBeginDocument{%
  \providecommand\BibTeX{{%
    \normalfont B\kern-0.5em{\scshape i\kern-0.25em b}\kern-0.8em\TeX}}%
}

%% file: abstract.tex
\abstract{%
Video generation is progressing beyond isolated clips toward long-form narratives and interactive worlds, requiring models to preserve identities, follow user controls, and remain stable over extended rollouts. We present \textbf{JoyAI-Echo-1.5}, a unified audio-visual generation system with two purpose-built variants. The \textbf{long-video} variant introduces composable cross-shot memory that aggregates visual evidence across multiple prior shots and speaker cues derived from speech-filtered full-shot audio, enabling persistent character appearance and voice identity across flexible combinations of text, image, and memory conditioning. The \textbf{world model} variant converts heterogeneous navigation inputs into calibrated metric 6-DoF camera trajectories and injects them through a geometry-aware conditioning pathway, enabling controller-agnostic interaction across flexible viewpoints. To support efficient long-horizon generation, we transform a bidirectional audio-visual backbone into a causal few-step generator using progressive teacher forcing and short- and long-horizon Self-Gradient Forcing on self-generated rollouts. Experiments demonstrate strong performance in both settings. JoyAI-Echo-1.5 achieves improvements over existing long-video baselines in cross-shot consistency, visual quality, text alignment, and speech fidelity. Its world-model variant ranks \textbf{first} on WBench, with an average score of 81.7, and achieves leading visual quality and long-horizon persistence on SANA-WM-Bench. Together, these results indicate that memory, geometric control, and rollout-aware training provide a practical foundation for generating coherent stories and continuously evolving interactive worlds.
}

%% file: sections/1_intro.tex
\section{Introduction}

The next frontier in video generation is not merely improving individual clips, but achieving \textbf{continuity and control.} Long-form creation requires characters and voices to remain consistent across shots, while interactive world modeling requires environments to respond precisely to user intent and remain stable during continuous rollouts. Yet existing video models are still largely designed around fixed temporal windows: they forget identities, drift when conditioned on self-generated histories, and struggle to translate heterogeneous action inputs into consistent motion. Overcoming these limitations requires more than scaling model capacity; it requires rethinking how models remember, act, and evolve over time.

We present \textbf{JoyAI-Echo-1.5}, a technical system comprising two purpose-built versions. Its \textbf{long-video version} targets long-form audio-visual generation and introduces composable cross-shot memory to preserve character appearance and speaker identity across changing scenes. It aggregates visual evidence from multiple prior shots, derives robust voice cues from speech-filtered full-shot audio, and supports flexible combinations of text, image, and memory conditions. Its \textbf{world-model version} targets interactive world modeling by converting heterogeneous navigation inputs into calibrated metric 6-DoF camera trajectories and injecting them through a geometry-aware control pathway. The first extends stories beyond individual shots; the second transforms generated video into an environment that can be explored.

Both versions are designed to operate beyond fixed temporal windows. JoyAI-Echo-1.5 progressively converts a bidirectional audio-visual backbone into a causal few-step generator through audio-visual teacher forcing followed by short- and long-horizon Self-Gradient Forcing on self-generated rollouts. Training on model-generated histories exposes the system to accumulated generation errors, helping it preserve synchronized audio-visual dynamics and sustain long streaming sequences within a bounded computational budget. \textbf{The long-video version gives generation memory. The world-model version gives it control. Together, JoyAI-Echo-1.5 pushes video generation from creating isolated moments toward sustaining stories and evolving worlds.}

\textbf{Main contributions.} Our main contributions are summarized as follows:
\begin{itemize}
    \item \textbf{A long-video system that remembers.} We introduce multi-shot audio-visual memory with speech-filtered voice representations, separated positional regions, and composable text, image, and memory conditioning to preserve identity across shots.

    \item \textbf{A world model that responds.} We establish a controller-agnostic action interface based on calibrated metric 6-DoF trajectories, enabling precise camera control across heterogeneous data sources, environments, and viewpoints.

    \item \textbf{A generator built to continue.} We develop a causal few-step training pipeline that combines audio-visual teacher forcing with short- and long-horizon Self-Gradient Forcing, allowing JoyAI-Echo-1.5 to learn from its own rollout distribution and remain stable over extended generation.
\end{itemize}

%% file: sections/2_Data.tex
\section{Data}

\subsection{Long-Video Training Data}
\label{sec:long_data}

\begin{figure*}[t]
    \centering
    \includegraphics[width=\textwidth]{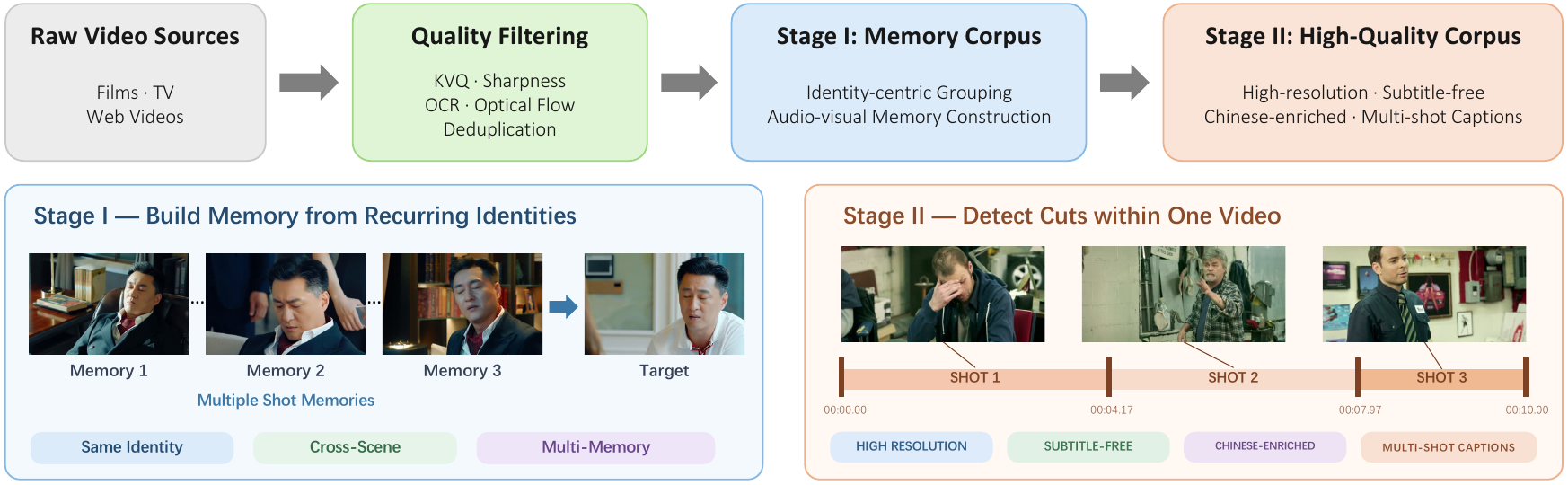}
    \caption{Two-stage data construction pipeline of JoyAI-Echo-1.5.
    Raw videos are quality-filtered and organized into an identity-centric
    Memory corpus and a high-quality corpus with broader resolution,
    language, and multi-shot coverage.}
    \label{fig:v15_data_pipeline}
\end{figure*}

JoyAI-Echo-1.5 adopts a two-stage data curriculum, as illustrated in
Fig.~\ref{fig:v15_data_pipeline}. \textbf{Stage I} reuses the identity-centric corpus of JoyAI-Echo-1.0~\citep{li2026joyai} and constructs memory--target pairs from scene-disjoint clips of the same recurring character to train audio-visual memory conditioning. \textbf{Stage II} uses high-quality videos with broader resolution and aspect-ratio coverage, increases the proportion of subtitle-free and Chinese-language samples, and incorporates captioned single-shot and multi-shot videos, with the latter explicitly describing transitions across cuts. It improves general audio-visual quality, spatial and linguistic coverage, and transition modeling, while preserving text-to-audio-video (T2AV) and image-to-audio-video (I2AV) generation capabilities. Both stages use the same quality-control pipeline, with their sampling distributions and caption formats adapted to their respective objectives.

\subsubsection{Stage-I Memory-Construction Corpus}
\label{sec:memory_data}

\paragraph{Identity-centric source groups.}
The memory-construction corpus retains the identity-centric organization introduced in JoyAI-Echo-1.0~\citep{li2026joyai}. Starting from long-form films, television episodes, and web videos, recurring characters are associated with scene-disjoint single-shot clips. Following the extraction pipeline of v1.0, the data undergo global identity discovery, scene grouping, local tracklet assignment, and quality filtering. Consequently, the basic unit used by v1.5 is not an isolated video clip, but an identity group
\begin{equation}
    \mathcal{G}_k=\{S_{k,n}\}_{n=1}^{N_k},
    \qquad
    S_{k,n}=(V_{k,n},A_{k,n}),
\end{equation}
where all clips in $\mathcal{G}_k$ contain the same recurring character while differing in scene, clothing, viewpoint, pose, motion, illumination, facial expression, dialogue, and acoustic environment. Maintaining this diversity is important because a useful memory representation should capture persistent character-level evidence rather than scene-specific appearance alone.

\paragraph{Memory-target sample construction.}
At each training iteration, one clip $S_{k,t}$ is selected as the prediction target and a variable-size subset of the remaining clips is sampled as memory:
\begin{equation}
    \mathcal{M}_{k,t}
    =\{S_{k,i_1},S_{k,i_2},\ldots,S_{k,i_K}\},
    \qquad i_j\neq t.
\end{equation}
The memory and target clips therefore show the same identity through different observations rather than through temporally adjacent frames from the same trajectory. This separation discourages direct content copying and requires the model to recover stable identity information under changes in background, camera pose, wardrobe, action, and speech content. The number of sampled memory shots $K$ is varied during training so that the model does not rely on a fixed memory capacity and remains effective when only a small amount of character evidence is available.

For every selected memory shot, one video frame is randomly sampled and encoded as visual memory. In contrast to the visual branch, which uses a compact frame-level observation, the complete shot audio is retained. The audio is first processed by the speech-filtering operator and is then encoded as audio memory. Using the complete speech-bearing interval provides a more reliable description of speaker characteristics than selecting a short response window, while speech filtering reduces the leakage of background music and environmental effects into the speaker condition. The visual and audio observations from each memory shot preserve the same slot order, allowing the model to jointly access the appearance and voice evidence associated with that shot.
\paragraph{Conditioning configurations.}
The availability of memory and the first-frame image condition is independently varied when constructing training samples. This produces four conditioning configurations under a unified interface:
\begin{itemize}
    \item \textbf{T2AV}: text-to-audio-video generation without memory or an input image;
    \item \textbf{I2AV}: audio-video generation conditioned on text and the first frame;
    \item \textbf{MT2AV}: text-to-audio-video generation conditioned on audio-visual memory;
    \item \textbf{MTI2AV}: audio-video generation jointly conditioned on memory, text, and the first frame.
\end{itemize}
The memory-construction corpus primarily supplies MT2AV and MTI2AV examples, while the same sample representation also supports T2AV and I2AV when the corresponding conditions are absent. Randomizing both memory length and image-condition availability prevents the model from binding a particular generation task to a fixed input layout. It also enables memory and the first-frame condition to be used compositionally, rather than treating memory-conditioned generation and image-conditioned generation as separate models.

\subsubsection{Stage-II High-Quality and Multi-Shot Enhancement Corpus}
\label{sec:hq_data}

The identity-organized corpus provides the supervision required for cross-shot memory retrieval, but its distribution is constrained by the availability of recurring characters and scene-disjoint observations. Training exclusively on such samples can reduce the diversity and standalone quality of ordinary T2AV and I2AV generation. Stage II therefore uses an enhancement corpus selected independently of the identity-group requirement. It combines high-quality single-shot videos with samples containing multiple shots and explicit transitions. Its purpose is to preserve and strengthen general audio-visual generation, broaden spatial and linguistic coverage, and improve transition modeling after the memory interface has been established in Stage I.

\paragraph{Resolution and aspect-ratio coverage.}
The high-quality corpus contains videos at $720{\times}720$, $1280{\times}720$, and $720{\times}1280$, corresponding to square ($1{:}1$), landscape ($16{:}9$), and portrait ($9{:}16$) formats. Rather than concentrating training on one canonical spatial shape, we explicitly balance these formats so that the model encounters comparable high-quality examples across common generation settings. This improves robustness to changes in token layout and spatial composition and allows the same model to serve cinematic landscape generation, portrait-oriented content, and square social-media formats.

\paragraph{Subtitle-free visual data.}
The retained samples emphasize clear visual content and stable audio-visual quality, with particular attention to videos without embedded subtitles. Burned-in subtitles create persistent, high-contrast patterns that are strongly correlated with dialogue. When such patterns are overrepresented, the model may reproduce incomplete glyphs, garbled character fragments, or subtitle-like regions even when no text is requested. Increasing the proportion of subtitle-free video weakens this undesirable correlation and provides cleaner supervision for faces, body motion, lower-frame image regions, and speaking scenes. OCR metadata is still retained for filtering and analysis, but samples with dominant or repeatedly occluding text are excluded from the enhancement corpus.

\paragraph{Chinese-language expansion.}
We additionally increase the proportion of Chinese-language data, including Chinese captions and dialogue-oriented audio-visual content. The broader language distribution supplies direct supervision for Chinese prompt following, spoken-content generation, and the alignment between visible speaking motion and Chinese dialogue. This adjustment is particularly important for long-form generation, where errors in linguistic conditioning or speaker behavior can accumulate across shots. The language expansion is performed together with visual-quality and subtitle filtering so that improved Chinese coverage does not come at the cost of image cleanliness or acoustic fidelity.

\subsubsection{Quality Filtering and Distribution Balancing}
\label{sec:v15_filter}

Both corpora inherit the multi-axis quality-control pipeline of JoyAI-Echo-1.0, but the retained distribution is recalibrated for high-resolution and multi-shot training.

\paragraph{Learned perceptual quality.}
We first apply a learned video-quality estimator~\citep{lu2024kvq} to assess perceptual quality over both spatial content and temporal behavior. It removes samples with severe compression artifacts, abnormal color, unstable exposure, visually distracting noise, or consistently low aesthetic quality. This learned score complements the following hand-crafted measurements: it captures degradations that are difficult to describe with one low-level statistic, whereas the explicit operators provide interpretable control over identity visibility, text overlays, and motion coverage.

\paragraph{Structural sharpness.}
Uniformly sampled frames are evaluated using the variance of the Laplacian. For a sampled frame $I_t$, its sharpness response is
\begin{equation}
    s_t^{\mathrm{sharp}}=\operatorname{Var}\!\left(\nabla^2 I_t\right),
\end{equation}
and the frame responses are robustly aggregated into a clip-level score
\begin{equation}
    S_{\mathrm{sharp}}(V)
    =\operatorname{Agg}\left(\{s_t^{\mathrm{sharp}}\}_{t\in\mathcal{T}}\right),
\end{equation}
where $\mathcal{T}$ denotes the uniformly sampled timestamps. Unlike face-detection confidence alone, this operator directly measures whether structural details remain visible. It is therefore particularly important for memory training: a face may still be detected under defocus, motion blur, or low spatial resolution, while providing unreliable identity supervision. Clips with persistently weak responses or insufficiently clear character observations are removed.

\paragraph{Text and overlay filtering.}
OCR and overlay detectors identify subtitles, logos, watermarks, large text regions, and burned-in graphics. We consider not only detected text content but also its occupied area, temporal persistence, and overlap with the main character. Clips in which text dominates the image or repeatedly occludes faces and bodies are rejected. For the high-quality enhancement corpus, this filter is made deliberately stricter to increase the proportion of subtitle-free data. OCR metadata is nevertheless preserved for caption construction: an empty result is explicitly verbalized as no on-screen text or subtitles, whereas visible text is introduced with an OCR marker.

\paragraph{Motion statistics.}
Dense optical flow~\citep{farneback2003optical} is used to estimate whole-frame dynamics. Given the flow field $\mathbf{u}_t(p)$ between adjacent frames, we compute
\begin{equation}
    M(V)=\frac{1}{T-1}\sum_{t=1}^{T-1}
    \frac{1}{|\Omega|}\sum_{p\in\Omega}
    \left\|\mathbf{u}_t(p)\right\|_2,
\end{equation}
where $\Omega$ is the image domain. The motion score is not used to retain only high-motion samples; instead, it supports distribution control. Very low-motion clips are downsampled to prevent the corpus from being dominated by static talking heads, while sharpness and perceptual quality filters prevent large flow caused by blur, camera shake, or corrupted frames from being mistaken for useful motion. The retained data therefore covers static dialogue, subtle facial and body motion, camera movement, and larger character displacement without collapsing toward either motion extreme.

\paragraph{Identity-level coverage and de-duplication.}
For the memory-construction corpus, all of the above filters are applied at both clip and identity-group levels. If an identity no longer contains enough scene-disjoint high-quality clips after filtering, the complete group is removed because it cannot support a meaningful memory--target split. Within each identity, perceptual hashes, color histograms, and DINOv2 background similarity are used to suppress near-duplicate observations. This ensures that adding memory slots introduces new evidence about appearance, viewpoint, scene, or voice rather than repeated copies of the same shot.

\paragraph{Distribution balancing.}
Finally, the retained samples are balanced across square, landscape, and portrait formats; Chinese and English content; subtitle presence; and motion ranges. The identity-organized and high-quality corpora are also mixed so that memory-conditioned examples do not overwhelm ordinary T2AV and I2AV training. These controls prevent one source domain or generation mode from dominating optimization and allow improvements in composable memory conditioning to be obtained without sacrificing general visual and acoustic fidelity.

\subsubsection{Single- and Multi-Shot Audio-Visual Captioning}
\label{sec:v15_caption}

JoyAI-Echo-1.5 expands the clip-level audio-visual annotation of v1.0 into bilingual captions covering both single-shot and multi-shot videos. For each structure, we construct a text-to-video caption and derive a corresponding image-to-video caption. The T2V caption must independently specify the visible subjects and their temporal behavior, whereas the I2V caption can rely on the input first frame for appearance information and is therefore rewritten to reduce redundant identity descriptions.

\paragraph{Single-shot T2V captions.}
For a single-shot clip, structured fields are concatenated in the following order:
\begin{equation}
    c_{\mathrm{single}}^{\mathrm{T2V}}
    =R\oplus A\oplus S\oplus C\oplus B\oplus O\oplus E\oplus G,
\end{equation}
where $R$ is \texttt{Roles\_and\_Subjects}, $A$ is \texttt{Action\_and\_Dialogue}, $S$ is \texttt{Style}, $C$ is \texttt{Camera\_Movement}, $B$ is \texttt{Background}, $O$ is the subtitle/OCR description, $E$ is \texttt{Sound\_Effects}, and $G$ is \texttt{BGM}. The action field describes actions, expressions, body movement, and dialogue in temporal order, while voice timbre and speaking manner are retained in the subject and dialogue descriptions. When no OCR or subtitle is detected, $O$ explicitly states that no on-screen text or subtitles are present; otherwise, the recognized content is introduced by an OCR marker. Repeated empty-text statements are removed during concatenation.

\paragraph{Multi-shot T2V captions.}
For a video containing $N$ shots, we use the rewritten identity-consistency annotation and serialize the individual shot records in chronological order:
\begin{equation}
    c_{\mathrm{multi}}^{\mathrm{T2V}}
    =H_N\oplus
    \bigoplus_{s=1}^{N}
    \left[Q_s\oplus R_s\oplus A_s\oplus C_s\oplus O_s\oplus E_s\oplus G_s\right],
\end{equation}
where $H_N$ states the number of shots and $Q_s$ is the explicit \texttt{Shot~$s$} boundary. Each shot preserves its own subjects, actions and dialogue, camera behavior, OCR state, sound effects, and background music. This format provides direct supervision for shot order, subject changes, dialogue allocation, camera changes, and soundtrack continuity across cuts. Characters recurring across shots retain consistent identifiers in the T2V annotation, while characters appearing for the first time are fully introduced in the shot where they enter.

\paragraph{I2V caption rewriting.}
Both single-shot and multi-shot T2V captions are converted into I2V captions. Because the first frame already provides the initial appearance, the full identifier-based subject description in the first shot is replaced by a concise semi-referential phrase, such as ``the short-haired man'' or ``the uniformed middle-aged man.'' The synthetic identifiers such as \texttt{ID\_A} are removed. Voice timbre, speaking manner, dialogue, sound effects, and background music are retained because they cannot be recovered reliably from the image. For multi-shot videos, any new character appearing after the first shot receives a sufficiently detailed natural-language description at first appearance, but is likewise described without an identifier token. The complete caption is then lightly shortened to remove repeated visual facts without discarding temporal actions or audio information.

A bilingual I2V instruction prefix is randomly sampled and prepended to the rewritten caption. The same procedure is applied independently to the Chinese and English annotations, and the resulting strings are stored in \texttt{video\_audio\_i2v\_caption}. This design reduces conflict between the image and redundant textual appearance descriptions while preserving the information required to generate motion, dialogue, voice, sound, and multi-shot evolution.

\begin{table*}[t]
\centering
\caption{Representative abbreviated examples of the four caption formats used in JoyAI-Echo-1.5. English renderings are shown for readability, while both Chinese and English captions are retained during training. T2V captions preserve explicit identity definitions, whereas I2V captions use concise natural references because appearance is supplied by the first frame.}
\label{tab:v15_caption_examples}
\footnotesize
\setlength{\tabcolsep}{4pt}
\renewcommand{\arraystretch}{1.18}
\begin{tabularx}{\textwidth}{p{1.45cm}XX}
\toprule
\textbf{Structure} & \textbf{T2V caption} & \textbf{I2V caption} \\
\midrule
\textbf{Single shot}
& \texttt{ID\_A} is a young man with short brown hair, wearing a dark gray hoodie, with a clear mid-range voice. He sits slightly off-camera, blinks naturally, and says, ``my games often focus on that ...'' A static chest-up close-up uses shallow depth of field in an indoor gym. No on-screen text or subtitles. No obvious sound effects. No background music.
& \textit{[Random I2V prefix]} A short-haired man speaks with a clear, stable mid-range voice and a natural conversational tone. He sits slightly off-camera, blinks naturally, and says, ``my games often focus on that ...'' A static chest-up close-up keeps him in focus against a softly blurred gym. No on-screen text or subtitles. No obvious sound effects. No background music. \\
\midrule
\textbf{Multi-shot}
& This video contains two shots. \textbf{Shot 1:} \texttt{ID\_A} is a middle-aged man in a dark green dress uniform with a deep voice; \texttt{ID\_B} is a middle-aged man in digital camouflage with a bright voice. They walk and talk side by side as a medium shot follows them. OCR: military slogans and a name label. Soft piano music plays. \textbf{Shot 2:} \texttt{ID\_C} is a silent middle-aged man in a dark blue police uniform. A police formation advances in a static medium shot. No OCR or subtitles; the piano music continues.
& \textit{[Random I2V prefix]} This video contains two shots. \textbf{Shot 1:} A middle-aged man in a dress uniform and a middle-aged man in camouflage walk and talk side by side. The former has a deep, steady voice and the latter a bright, natural voice. A medium shot follows them; soft piano music plays. \textbf{Shot 2:} A serious middle-aged man in a dark blue police uniform appears silently within an advancing formation. A static medium shot follows the group. No OCR or subtitles; the piano music continues. \\
\bottomrule
\end{tabularx}
\end{table*}

The two-stage curriculum separates the acquisition of memory conditioning from the subsequent enhancement of general quality and transition modeling. Stage I teaches the model to retrieve and compose cross-shot identity evidence. Stage II uses high-quality subtitle-free data with increased Chinese coverage to maintain standalone T2AV and I2AV fidelity; within the same stage, captioned single-shot samples preserve fine-grained intra-shot actions and multi-shot samples provide explicit supervision for transitions across cuts. This progression allows JoyAI-Echo-1.5 to extend long-form controllability and composable memory conditioning while maintaining broad spatial, linguistic, visual, and acoustic coverage.

\subsection{Action Annotation and Trajectory Data}

Action-conditioned world generation requires a consistent motion representation across data collected from different sources. However, these sources provide substantially different levels of supervision: synthetic UE renders expose exact camera poses and control inputs, internal gameplay provides action logs but not camera poses, while human gameplay and web videos provide audio-visual observations but no explicit action logs or camera telemetry. We therefore build a unified processing pipeline that converts all sources into metric 6-DoF camera trajectories and, whenever available, aligns them with source-native control signals.

\noindent\textbf{Trajectory Data from Heterogeneous Sources.}
We collect motion supervision from three complementary data sources:

\begin{itemize}[leftmargin=*,itemsep=4pt,topsep=3pt]
\item \textit{Unreal Engine (UE) Renders.}
UE environments provide accurate geometric supervision. Avatars are controlled through local navigation commands under the engine's native physics and collision system. Each rollout is rendered from one first-person and four character-relative third-person cameras. For every frame, we synchronously record RGB observations, keyboard inputs, camera intrinsics, and ground-truth camera-to-world poses in metric units. During training, one viewpoint is randomly sampled from each sequence, encouraging the model to generalize across camera configurations.

\item \textit{Internal Gameplay.}
Internal game environments provide diverse interaction scenarios, including open-world navigation, driving, and mounted traversal in both first- and third-person views. Scripted agents execute predefined control sequences, allowing us to record accurately aligned action commands together with RGB video, stereo audio, and scene metadata. Interface elements such as HUDs and crosshairs are removed during rendering. Since these game engines do not expose camera pose matrices, their 6-DoF trajectories are estimated from the recorded videos and temporally aligned with the corresponding action logs.

\item \textit{Human Gameplay and Web Video.}
These videos complement scripted data with natural navigation behavior, human reaction patterns, and richer audio-visual content. They do not contain explicit action logs or camera telemetry, so we use the recovered camera trajectory itself as motion supervision. We extract temporally continuous, shot-consistent segments of approximately one minute and estimate their metric 6-DoF camera motion using geometry-based reconstruction.

\end{itemize}

Together, these sources provide a spectrum from precise synthetic supervision to large-scale natural motion, while sharing the same trajectory representation.

\noindent\textbf{Long-Sequence Metric Pose Recovery.}
For videos without ground-truth camera poses, directly estimating motion on short training clips often leads to scale ambiguity and unstable trajectories. We instead estimate camera motion over longer, continuous windows of approximately one minute and only afterward divide them into training clips.

Specifically, we apply the metric-depth-assisted ViPE pipeline~\citep{huang2025vipe} to temporally subsampled frames. The longer temporal context provides a larger geometric baseline and improves the robustness of pose reconstruction. We then recover frame-rate trajectories by interpolating rotations with Spherical Linear Interpolation (SLERP) and translations with linear interpolation. Finally, both estimated and ground-truth trajectories are transformed into a common camera-to-world coordinate convention, metric scale, axis orientation, and temporal sampling rate.

\noindent\textbf{Trajectory Quality Filtering.}
We apply trajectory-level filtering before training. Clips are removed when pose estimation has low confidence or exhibits implausible motion, including severe frame-to-frame jitter, abrupt rotational changes, negligible displacement, or unstable camera intrinsics. Ground-truth UE trajectories are subjected to the same motion-range checks, while their original metric poses are kept unchanged.

\noindent\textbf{Action-Decoupled Text Annotation.}
We structure each audio-visual caption into persistent scene attributes, a dynamic-motion narrative, and audio-related descriptions. Persistent attributes include the scene, visual style, viewpoint, and subject identity, while the audio component describes speech and environmental sounds. During Action-SFT, we omit both the dynamic-motion narrative and the audio-related fields, retaining only the persistent static descriptors because this stage uses a visual-only denoising objective. During Joint-FT, we restore the audio-related fields while continuing to exclude the dynamic-motion narrative. In both stages, camera motion is specified explicitly by the trajectory condition rather than redundantly through language. This encourages the model to follow the provided trajectory while preserving the remaining prompt semantics.

%% file: sections/3_echolong.tex


\section{Long-Horizon Video Generation}
\label{sec:memory_v15}
\begin{figure*}[t]
    \centering
    \includegraphics[width=\textwidth]{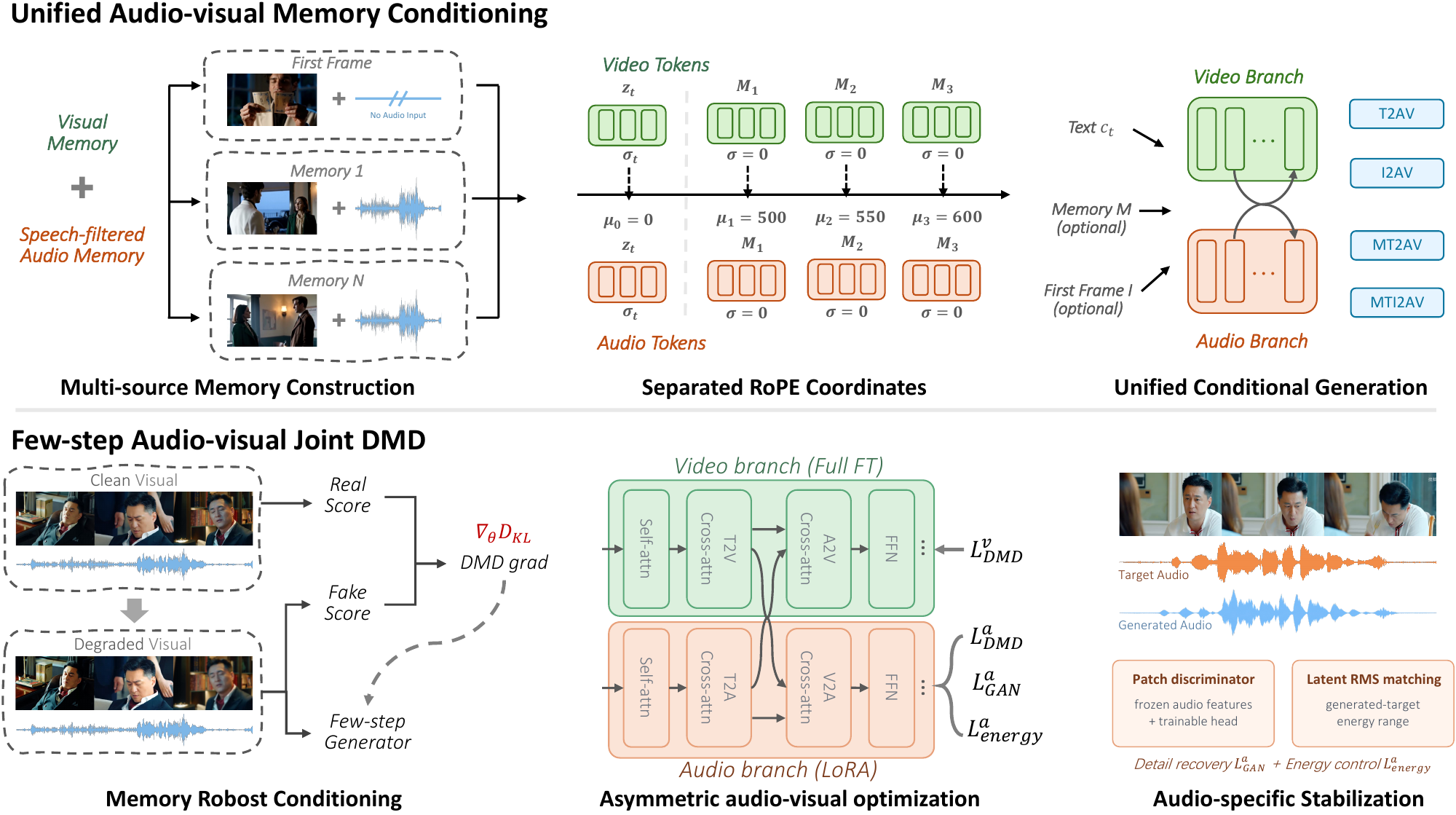}
    \caption{Overview of memory-conditioned long-form generation and few-step acceleration in JoyAI-Echo-1.5.
Top: unified conditioning with single-frame visual memories, speech-filtered audio memories, and separated RoPE coordinates supports multiple audio-visual generation tasks.
Bottom: memory-robust joint DMD combines asymmetric video--audio optimization with audio-specific adversarial and energy constraints.}
    \label{fig:v15_memory_dmd}
\end{figure*}
JoyAI-Echo-1.0~\citep{li2026joyai} introduced a long-term audio-visual memory mechanism that reuses compact visual and acoustic cues from previous shots, substantially improving character appearance and speaker-timbre consistency during multi-shot generation. Building on this foundation, JoyAI-Echo-1.5 improves both long-range consistency and generation efficiency. First, it introduces a unified audio-visual memory-conditioning framework that constructs visual memory from multiple shots, extracts audio memory from speech-filtered full-shot audio, separates memory and target tokens through RoPE coordinates, and jointly trains different combinations of memory and image conditions. Second, it applies distribution matching distillation to convert the original multi-step generator into an 8-step student, with optimization strategies designed for the distinct characteristics of video and audio generation. We first formulate the long-horizon generation task and present the unified memory-conditioning framework, and then describe the proposed distillation method for efficient audio-visual generation.

\subsection{Unified Audio-Visual Memory Conditioning}
\label{sec:unified_memory}

\paragraph{Task formulation.}

Given a story script represented by shot-level text conditions
$C=\{c_t\}_{t=1}^{T}$, JoyAI-Echo generates a sequence of synchronized
audio-visual shots $S=\{S_t\}_{t=1}^{T}$, where $S_t=(V_t,A_t)$. Let $M_0$
denote an optional memory bank supplied before generation, and let
$I=\{I_t\}_{t=1}^{T}$ denote optional image conditions for individual shots.
Long-form generation is factorized as:
\begin{equation}
  p_\theta(S\mid C,M_0,I)
  =\prod_{t=1}^{T}p_\theta(S_t\mid c_t,M_{t-1},I_t),
  \qquad
  M_t=\mathcal{U}(M_{t-1},S_t).
  \label{eq:v15_factorization}
\end{equation}
Before generating shot $t$, $M_{t-1}$ contains both the initial memory $M_0$
and the memory extracted from shots $S_1,\ldots,S_{t-1}$. After $S_t$ is
generated, the update function $\mathcal{U}$ extracts its audio-visual
information and updates the memory bank for the next shot. The image condition
$I_t$, when provided, specifies the first frame of the current target shot and
does not consume a memory slot.

\paragraph{Audio-visual memory construction.}
Each memory item is written as $m_i=(m_i^v,m_i^a)$, where $m_i^v$ and $m_i^a$
denote its visual and audio components. From the available history, we select
up to $K$ shots and arrange them in chronological order. For the $i$-th
selected shot, one frame is sampled and encoded by the video VAE:
\begin{equation}
  m_i^v=E_v\!\left(V_i[f_i]\right),
  \qquad
  f_i\sim\mathcal{U}\{1,\ldots,|V_i|\}.
  \label{eq:v15_visual_memory}
\end{equation}
The resulting latents provide compact visual observations from different
historical shots instead of restricting the condition to a single adjacent
frame.

For audio memory, we use the complete audio content of each selected shot.
Since full-shot audio may contain background music, environmental sounds, and
other non-speech components, we first apply a speech-filtering operator
$\mathcal{F}_{\mathrm{speech}}$ and then encode the filtered waveform using the
audio VAE:
\begin{equation}
  \widetilde{A}_i=
  \mathcal{F}_{\mathrm{speech}}(A_i),
  \qquad
  m_i^a=E_a(\widetilde{A}_i).
  \label{eq:v15_audio_memory}
\end{equation}
Here, $\mathcal{F}_{\mathrm{speech}}$ is a generic speech-separation operator
that can be instantiated by different filtering models; our current
implementation uses a speech-extraction model through the
Music-Source-Separation-Training (MSST) toolkit. In
v1.0, a short audio window is selected according
to acoustic energy. However, the maximum-response window may be dominated by
music or other loud background components, which can be repeatedly reinforced
when generated audio is reused during autoregressive rollout. The new
construction instead uses the complete filtered speech content of each shot,
providing cleaner and more stable speaker information. The visual and audio
latents are then arranged in the same slot order, so that each memory item
contains one compact visual observation and its corresponding event-level
audio segment.

\paragraph{RoPE for memory conditioning.}
All historical conditions and user-provided memory inputs are represented in
a common memory format. Following the discontinuous RoPE offsets introduced in
ShotStream~\citep{luo2026shotstream}, we assign separated temporal RoPE regions
to the current target, historical memory, and memory supplied before generation.
The target video and audio tokens retain their original temporally aligned
coordinates in a low-offset region below 500. When an image condition $I_t$ is
provided, its tokens remain in the coordinate system of the current target shot
and are placed at temporal coordinate zero as its clean first frame.

The visual and audio components extracted from the $i$-th historical shot share
the same temporal RoPE center:
\begin{equation}
  \mu_i=500+50(i-1),
  \label{eq:v15_memory_position}
\end{equation}
such that successive audio-visual memory items are centered at
$500,550,600,\ldots$. All spatial tokens of the sampled visual memory frame use
$\mu_i$ as their temporal coordinate while retaining their original spatial
RoPE coordinates. For the corresponding audio latent with native temporal
coordinates $\{\tau_{i,\ell}^a\}_{\ell=1}^{L_i^a}$, we assign:
\begin{equation}
  r_{i,\ell}^{a}
  =
  \mu_i+\tau_{i,\ell}^{a}-\bar{\tau}_i^{a},
  \qquad
  \bar{\tau}_i^{a}
  =
  \frac{1}{L_i^a}
  \sum_{\ell=1}^{L_i^a}\tau_{i,\ell}^{a},
  \label{eq:v15_audio_memory_position}
\end{equation}
where $\ell$ indexes audio tokens. In other words, the temporal RoPE
coordinates of the complete audio latent are shifted together so that its
midpoint is aligned with the same center $\mu_i$. This operation preserves the
relative temporal structure of the audio event and does not modify the audio
latent features themselves.

Memory supplied before generation is encoded independently and placed in a
high-offset region:
\begin{equation}
  M_0=\{m_j^0\}_{j=1}^{J},
  \qquad
  m_j^0=(m_j^{0,v},m_j^{0,a}),
  \label{eq:v15_memory_rope}
\end{equation}
where
\begin{equation}
  m_j^{0,v}=E_v(X_j^{\mathrm{mem}}),
  \qquad
  m_j^{0,a}=E_a\!\left(
    \mathcal{F}_{\mathrm{speech}}(A_j^{\mathrm{mem}})
  \right),
  \qquad
  \rho_j\geq5000.
  \label{eq:v15_initial_memory}
\end{equation}
The visual and audio components of the $j$-th initial memory item are both
centered at $\rho_j$ using the same positional assignment: the visual tokens
use $\rho_j$ as their temporal coordinate, while the audio RoPE coordinates
are shifted together so that their midpoint is aligned with $\rho_j$.
The centers $\{\rho_j\}_{j=1}^{J}$ are distinct and are assigned with enough
spacing to keep the shifted audio-coordinate spans of different memory items
non-overlapping.
Consequently, the current target, historical memory, and initial memory occupy
separated temporal RoPE regions. This separation keeps memory outside the
target-shot timeline while allowing memory and image conditions to be provided
simultaneously.

\paragraph{Memory interaction.}
After memory construction and RoPE assignment, memory and target tokens are
jointly processed by the model. Unlike v1.0, v1.5 removes the memory-specific
audio self-attention mask and the slot-paired cross-modal attention mask. All
valid tokens interact through the original self-attention and bidirectional
cross-modal attention paths. Their roles remain distinguishable because memory
tokens use dedicated RoPE coordinates, remain clean, and are excluded from the
prediction loss, whereas target tokens follow the target timeline and are
optimized as generation outputs. This unified interaction removes additional
hard constraints and naturally supports the joint use of memory and image
conditions.

\paragraph{Joint conditional training.}
Training follows the two-stage data curriculum described in
Sec.~\ref{sec:long_data}. Throughout the curriculum, we use a unified conditional
interface that supports text-to-audio-video (T2AV),
image-to-audio-video (I2AV), memory-conditioned text-to-audio-video (MT2AV),
and memory-conditioned text-and-image-to-audio-video (MTI2AV). Let
$b_M,b_I\in\{0,1\}$ indicate whether audio-visual memory and a first-frame
image condition are provided. The model condition for shot $t$ is:
\begin{equation}
  \mathcal{C}_t=(c_t,\,b_M M_{t-1},\,b_I I_t),
  \label{eq:v15_joint_condition}
\end{equation}
where $(b_M,b_I)=(0,0),(0,1),(1,0),(1,1)$ correspond to T2AV, I2AV,
MT2AV, and MTI2AV, respectively.

Let $z_0^v$ and $z_0^a$ denote the clean target video and audio latents, and
let $\epsilon^v,\epsilon^a\sim\mathcal{N}(0,\mathbf{I})$. For each modality
$q\in\{v,a\}$, rectified-flow training constructs the interpolation and target
velocity for the tokens to be predicted:
\begin{equation}
  z_\sigma^q=(1-\sigma)z_0^q+\sigma\epsilon^q,
  \qquad
  u^q=\epsilon^q-z_0^q.
  \label{eq:v15_flow_target}
\end{equation}
We write the joint noised audio-visual target as
$z_\sigma=(z_\sigma^v,z_\sigma^a)$ and the paired noise as
$\epsilon=(\epsilon^v,\epsilon^a)$. Memory tokens are inserted as clean
conditions with zero noise level and are excluded from the prediction loss.
When $b_I=1$, the first target video frame is also kept clean and excluded from
the video loss. Let $W_v$ and $W_a$ be binary loss masks selecting only noised
target tokens. The joint training objective is:
\begin{equation}
  \mathcal{L}
  =\mathbb{E}_{\mathrm{data},\epsilon,\sigma,b_M,b_I}\!\left[
    \left\|W_v\odot
      \left(v_\theta^v(z_\sigma,\mathcal{C}_t)-u^v\right)
    \right\|_2^2
    +\lambda_a
    \left\|W_a\odot
      \left(v_\theta^a(z_\sigma,\mathcal{C}_t)-u^a\right)
    \right\|_2^2
  \right],
  \label{eq:v15_joint_training}
\end{equation}
where $\lambda_a$ balances the audio and video objectives. The resulting
training distribution covers the four combinations of $(b_M,b_I)$, enabling
generation without memory, with image conditioning alone, with memory alone,
and with both conditions. This unified formulation supports T2AV, I2AV,
MT2AV, and MTI2AV without changing the model interface or introducing
separate objectives.

\subsection{Acceleration with Distribution Matching Distillation}
\label{sec:dmd}

Long-form audio-visual generation requires both long-range consistency and low-latency inference. To accelerate JoyAI-Echo-1.5, we apply distribution matching distillation (DMD) to convert the original multi-step audio-visual generator into an 8-step student generator. 

\paragraph{Data curation and training pipeline.}
\label{sec:dmd_data_pipeline}

JoyAI-Echo-1.5 performs DMD using memory conditions constructed from real long-form videos following Sec.~\ref{sec:unified_memory}. From the large-scale training corpus, we curate a compact subset with high-quality audio-visual memory and diverse generation conditions. The resulting data mixture covers T2AV, I2AV, and MT2AV, together with samples annotated by multi-shot captions. During training, we randomly sample these tasks according to predefined mixing ratios, allowing the student generator to preserve both single-shot generation quality and memory-conditioned multi-shot capability.

We find that DMD is particularly sensitive to the quality and distribution of visual memory. Low-level discrepancies in memory appearance, including brightness and saturation shifts, can bias the conditional distribution learned by the student and become progressively amplified when generated shots are written back into memory. Before DMD training, we therefore calibrate the low-level statistics of visual memory to better match the target training distribution. This preprocessing substantially improves distillation stability and reduces quality drift during multi-shot rollout.

The curated task mixture and calibrated memory conditions are used throughout the complete DMD pipeline. We first perform high-shift DMD with a relatively large learning rate to establish video quality and global structure, and then continue from the converged checkpoint with a lower shift and learning rate to refine audio and other fine-grained details. In both stages, the video branch is fully fine-tuned, while the audio and cross-modal modules are adapted through LoRA.

\paragraph{Asymmetric audio-visual optimization.}
\label{sec:base_dmd}

Video and audio exhibit substantially different structures in the latent space. Video latents are highly structured and spatially redundant, with strong correlations across neighboring spatial locations and adjacent frames. Their temporal evolution is also relatively smooth, allowing global appearance, scene layout, and motion patterns to be represented through slowly varying latent structures. Audio latents, in contrast, encode dense one-dimensional temporal signals with rapidly changing local patterns. Fine-grained information such as phonetic content, pitch, transients, and acoustic textures may vary significantly within a short temporal interval~\citep{su2026omniforcing}.

These differences lead to asymmetric optimization requirements during few-step DMD. The video branch must substantially adapt its spatial and temporal representations to recover visual structure and motion along the shortened sampling trajectory. Restricting its trainable capacity results in insufficient adaptation and degraded visual fidelity. The audio branch, however, is more sensitive to aggressive distribution-matching updates. Small deviations in the audio latent can remove perceptually important local details, while excessive parameter updates may overwrite the acoustic prior of the multi-step teacher and introduce muffled, metallic, or electronic-sounding artifacts~\citep{tian2026audioxturbo}.

We therefore allocate different trainable capacities to the two modalities:
\begin{equation}
\theta_G^{\mathrm{train}}
=
\left\{
\theta_v^{\mathrm{full}},
\Delta\theta_a^{\mathrm{LoRA}},
\Delta\theta_{\mathrm{cross}}^{\mathrm{LoRA}}
\right\},
\label{eq:asymmetric_parameterization}
\end{equation}
where the video branch is fully fine-tuned, while the audio branch and cross-modal attention modules are adapted through low-rank adaptation (LoRA). Full fine-tuning provides sufficient capacity for the video branch to learn the structural and motion changes required by few-step generation. In contrast, LoRA constrains the effective update space of the audio branch, slowing its deviation from the well-trained acoustic prior. Applying LoRA to the cross-modal attention modules preserves the flexibility required to adapt audio-visual correspondence without aggressively modifying the audio representation itself. This asymmetric parameterization therefore balances video adaptability with audio stability during joint DMD training.

\paragraph{Coarse-to-fine DMD noise scheduling.}
\label{sec:two_stage_dmd}

In DMD, the clean prediction produced by the student generator is re-noised to an auxiliary noise level $\sigma$, at which the teacher and fake-score model estimate the real and generated scores~\citep{yin2024onestep,yin2024improved}. The choice of $\sigma$ is important because it determines the granularity of the distributional information exposed to score matching. Higher noise levels place greater emphasis on coarse distributional properties and help stabilize global structure, whereas lower noise levels retain more sample-specific information and provide stronger supervision for fine-grained details.

Video and audio benefit differently from these noise regimes. Video generation relies strongly on high-noise DMD updates to establish stable appearance, scene layout, motion, and temporal structure under the shortened sampling trajectory. Audio quality, in contrast, is particularly sensitive to the low-noise regime, where local temporal patterns such as phonetic details, pitch, transients, and acoustic textures are more directly preserved and corrected. A fixed noise regime therefore cannot simultaneously provide the most suitable optimization signal for both modalities.

We address this mismatch with a two-stage DMD schedule. The first stage uses a relatively large timestep shift and learning rate, biasing score matching toward higher noise levels to establish stable video quality and structure. Starting from the converged first-stage checkpoint, the second stage reduces both the timestep shift and learning rate, shifting the optimization toward lower noise levels to refine audio quality and other fine-grained details. This coarse-to-fine schedule accommodates the different noise-level preferences of video and audio during joint few-step distillation.

\paragraph{Memory-robust and task-adaptive DMD.}
\label{sec:memory_robust_dmd}

For the DMD stage described here, we consider T2AV/\allowbreak{}I2AV/\allowbreak{}MT2AV, where MT2AV generates the current shot conditioned on the audio-visual memory bank $M_{t-1}$. Although the training memory is constructed from real long-form videos, the memory available during inference is progressively extracted from previously generated shots. Even small generation errors can therefore alter the memory distribution and accumulate over a multi-shot rollout. A student trained only with undegraded memory may become sensitive to this discrepancy, leading to gradual degradation of visual quality and cross-shot consistency.

To improve robustness, we apply stochastic degradation to the visual memory during DMD training. The student generator and fake-score model receive the same probabilistically degraded memory, allowing the fake-score model to track the student distribution under imperfect memory conditions. In contrast, the real-score model receives the original, undegraded memory and provides a stable target corresponding to the desired conditional distribution. The resulting score difference encourages the student to recover a high-quality current shot even when its historical memory contains moderate errors. This stochastic degradation complements the visual-memory distribution calibration described above: calibration corrects systematic low-level biases in the training memory, while degradation simulates the sample-dependent errors encountered during autoregressive rollout.

Moreover, MT2AV and other tasks also exhibit different audio optimization dynamics. In T2AV or I2AV, the student must construct the complete audio distribution from text and noise withour any audio reference. In MT2AV, the memory condition already provides informative cues about speaker timbre, acoustic context, and temporal continuity, making its audio component easier to learn. Consequently, using the same audio weight for both tasks can cause the MT2AV audio objective to contribute disproportionately to joint DMD optimization.

We therefore introduce a task-dependent audio weight:
\begin{equation}
\mathcal{L}_{\mathrm{DMD}}
=
\mathcal{L}_{\mathrm{DMD}}^{v}
+
\lambda_a(\kappa)\mathcal{L}_{\mathrm{DMD}}^{a},
\label{eq:task_adaptive_dmd_objective}
\end{equation}
where $\kappa\in\{\mathrm{T2AV},\mathrm{I2AV},\mathrm{MT2AV}\}$ denotes the DMD task, with
$\lambda_a(\mathrm{T2AV/I2AV})=0.25$ and
$\lambda_a(\mathrm{MT2AV})=0.10$. The reduced MT2AV weight prevents its easier memory-conditioned audio objective from dominating joint distillation, while the larger weight retains sufficient supervision for learning audio generation directly from noise. Together, stochastic memory degradation and task-adaptive weighting improve the robustness of few-step generation across both single-shot and memory-conditioned settings.

\paragraph{Audio distillation with adversarial and energy constraints.}
\label{sec:audio_stabilization}

Despite the conservative Audio-LoRA parameterization and low-noise refinement stage, applying score-based DMD to audio remains challenging. The distribution-matching gradient mainly corrects the discrepancy between the teacher and student distributions at noisy latent states. For dense audio signals, this correction can favor dominant acoustic modes while suppressing subtle temporal variations, producing over-smoothed, muffled, or electronic-sounding outputs. We further observe that DMD may cause the overall audio latent level to increase continuously during training, leading to excessive loudness, clipping, and harsh distortion after decoding. We therefore introduce two complementary audio-specific objectives to constrain global signal level and local acoustic realism.

First, we match the root-mean-square (RMS) level of the generated audio latent to that of its ground-truth counterpart:
\begin{equation}
\mathcal{L}_{\mathrm{energy}}^{a}
=
\psi\left(
\log\left(\operatorname{RMS}(\hat{z}_0^a)+\varepsilon_{\mathrm{num}}\right)
-
\log\left(\operatorname{RMS}(z_0^a)+\varepsilon_{\mathrm{num}}\right)
\right),
\label{eq:audio_energy_constraint}
\end{equation}
where $\hat{z}_0^a$ and $z_0^a$ denote the predicted and ground-truth clean audio latents, respectively. The robust penalty $\psi$ limits the influence of extreme samples, and $\varepsilon_{\mathrm{num}}>0$ ensures numerical stability. Computing the discrepancy in the log domain makes the loss sensitive to relative rather than absolute level differences. This constraint suppresses persistent latent-level drift and reduces the risk of clipping after audio decoding.

Second, following AudioX-Turbo~\citep{tian2026audioxturbo}, we introduce a diffusion-based audio discriminator. We reuse intermediate features from the frozen audio teacher as the discriminator backbone and train only lightweight heads to distinguish generated and real audio latents. By preserving the pretrained audio representation and limiting adversarial training to the added heads, the discriminator provides a stable perceptual signal without modifying the teacher. Its local adversarial feedback complements the distribution-level DMD objective by preserving transients, acoustic textures, and other fine-grained temporal details.

The complete student-generator objective is
\begin{equation}
\begin{aligned}
\mathcal{L}_G
={}&
\mathcal{L}_{\mathrm{DMD}}^{v}
+
\lambda_a(\kappa)\mathcal{L}_{\mathrm{DMD}}^{a} 
&+
\lambda_{\mathrm{GAN}}\mathcal{L}_{\mathrm{GAN}}^{a}
+
\lambda_{\mathrm{energy}}\mathcal{L}_{\mathrm{energy}}^{a}.
\end{aligned}
\label{eq:complete_dmd_objective}
\end{equation}
The DMD terms align the global audio-visual generation distribution, the adversarial loss restores local acoustic details, and the RMS-level constraint stabilizes the overall audio signal level. Together, these objectives mitigate both perceptual over-smoothing and excessive energy growth during few-step distillation.

%% file: sections/4_wm.tex
\section{Action-Conditioned World Modeling}
\label{sec:method}

\subsection{Overview and Problem Formulation}

Given an optional clean media context $\mathcal{M}$, a text condition $\mathcal{C}$, and a user control sequence $U_{1:T}$, our goal is to jointly generate synchronized video $V_{1:T}$ and audio $A_{1:T}$ under user-specified camera intent. 
The media context $\mathcal{M}$ accommodates an empty condition $\varnothing$, a single reference image $I_{\mathrm{ref}}$, or an audio-visual prefix $(V_{1:\tau}, A_{1:\tau})$. 
We serialize user navigation inputs into a time-indexed metric camera trajectory event stream $\mathcal{E}_{P}=\{(t,P_t)\}_{t=1}^{T}$, where $P_t$ represents a calibrated relative 6-DoF pose.
The joint generation process is formulated as:
\begin{equation}
    p_{\theta,\phi}\!\left(V_{1:T}, A_{1:T} \mid \mathcal{M}, \mathcal{C}, \mathcal{E}_{P}\right),
    \label{eq:world-formulation}
\end{equation}
where $\theta$ denotes the audio-visual backbone, and $\phi$ represents the parallel relative trajectory-conditioning pathway. Static text fields specify scene semantics, while $\mathcal{E}_{P}$ dictates the continuous temporal trajectory evolution.
When $\mathcal{M}$ contains an observed audio-visual prefix, those prefix
samples are clamped and only the unobserved suffix is generated; the
full-sequence notation in Eq.~\eqref{eq:world-formulation} is used for
compactness.

\subsection{Unified Camera-Intent Interface \& Scale Calibration}
\label{sec:camera-intent-interface}

Realizing continuous world modeling across heterogeneous datasets requires overcoming two key obstacles: the incompatibility of controller-specific action spaces and the scale mismatch of relative motion amplitudes. To address these issues, we decouple user-facing controls from model-level conditions through a two-fold strategy: (1) mapping arbitrary navigation inputs into a unified, subject-agnostic 6-DoF trajectory, and (2) calibrating relative translations with a robust global scale.

\paragraph{Relative 6-DoF Representation.}
We define camera intent using relative 6-DoF trajectories rather than actor-specific control states. Let $T_t \in \mathrm{SE}(3)$ denote the camera-to-world extrinsic matrix at frame $t$. The relative motion between frames $i$ and $j$ is given by:
\begin{equation}
    \Delta T_{ij} = T_i^{-1}T_j, \qquad \boldsymbol{\xi}_{ij} = \operatorname{Log}(\Delta T_{ij})^\vee \in \mathbb{R}^{6},
\end{equation}
where $\boldsymbol{\xi}_{ij}$ is the corresponding 6-DoF twist coordinate. For discrete keyboard inputs $\mathbf{u}_t$, a fixed control mapping $\mathcal{F}_{\mathrm{ctrl}}$ converts each input into a step-wise camera increment:
\begin{equation}
    \boldsymbol{\xi}_t = \mathcal{F}_{\mathrm{ctrl}}(\mathbf{u}_t), \qquad T_t = T_{t-1}\operatorname{Exp}(\widehat{\boldsymbol{\xi}}_t), \qquad \Delta T_t = T_0^{-1}T_t,
\end{equation}
where $\widehat{\boldsymbol{\xi}}_t \in \mathfrak{se}(3)$. Continuous metric trajectories bypass $\mathcal{F}_{\mathrm{ctrl}}$ and directly produce relative poses $\Delta T_{1:T}$. This unifies raw actions and estimated poses under a single geometric interface.

\paragraph{Global Translation Scale Calibration.}
To ensure consistent velocity semantics without introducing boundary artifacts
in multi-turn rollouts, we calibrate relative translations across all data
sources using a fixed global dataset-level scale $s_{\mathrm{global}}$.
For clip $i$, let $\Delta\mathbf{t}_{i,k}$ denote the relative translation at
frame $k$ with respect to the initial frame. We compute
\begin{equation}
    s_{\mathrm{global}}
    =
    Q_{0.9}\!\left(
        \left\{
            \max_k \|\Delta\mathbf{t}_{i,k}\|_2
        \right\}_{i\in\mathcal{D}_{\mathrm{train}}}
    \right).
\end{equation}
Relative translations are calibrated via $\widehat{\mathbf{t}}_{i,k} = \Delta\mathbf{t}_{i,k} / s_{\mathrm{global}}$, yielding $P_{i,k} = [\Delta\mathbf{R}_{i,k} \mid \widehat{\mathbf{t}}_{i,k}]$ while keeping rotation matrices $\Delta\mathbf{R}_{i,k}$ intact.

\subsection{Relative Trajectory Conditioning}
\label{sec:camera-conditioning}

Instead of absolute Pl\"ucker coordinates which depend on arbitrary world origins, we inject trajectory geometry through Unified Camera Positional Encoding (UCPE)~\citep{zhang2025ucpe}. UCPE makes pairwise camera geometry invariant to global coordinate shifts while preserving calibrated relative translation magnitudes.

For a latent token $i=(t,s)$ at frame $t$ and spatial cell $s$, we define its local ray transformation $\mathbf{D}_i = (\mathbf{T}^{\mathrm{wr}}_i)^{-1}$ derived from calibrated pose $P_t$ and intrinsics $K_t$. We inject this geometry into parallel attention heads via matrix scaling $\mathbf{G}_i = \mathbf{I}_{d/8} \otimes \mathbf{D}_i$ combined with spatiotemporal RoPE:
\begin{equation}
\begin{aligned}
    \widetilde{\mathbf{q}}^{c}_i &= (\mathbf{G}_i^\top \oplus \operatorname{RoPE}_i)\mathbf{q}^{c}_i, \\
    (\widetilde{\mathbf{k}}^{c}_i, \widetilde{\mathbf{v}}^{c}_i) &= (\mathbf{G}_i^{-1} \oplus \operatorname{RoPE}_i)(\mathbf{k}^{c}_i, \mathbf{v}^{c}_i), \\
    \mathbf{o}^{c}_i &= (\mathbf{G}_i \oplus \operatorname{RoPE}_i^{-1}) \operatorname{Attn}_{\mathrm{cam}}(\widetilde{\mathbf{q}}^{c}, \widetilde{\mathbf{k}}^{c}, \widetilde{\mathbf{v}}^{c})_i.
\end{aligned}
\end{equation}
Here, $\mathbf{T}^{\mathrm{wr}}_i$ is the UCPE ray-to-world transform
constructed from $(P_t,K_t,s)$, so $\mathbf{D}_i$ maps world coordinates to
the ray-local frame; $d$ is the attention-head width, and the
superscript $c$ identifies the parallel camera-conditioning stream. The symbol
$\oplus$ denotes a block-diagonal direct sum with the spatiotemporal RoPE
operator, and $\operatorname{Attn}_{\mathrm{cam}}$ denotes attention in the
camera pathway. The resulting camera-attention outputs pass through
zero-initialized linear projections in the trajectory pathway parameterized by
$\phi$ and are added to the main video self-attention blocks. The trajectory
condition is applied exclusively to the visual backbone.
One UCPE pathway is shared across first- and third-person data.
Geometry alone does not determine whether a trajectory is realized from an egocentric or third-person observer--subject configuration, so the static viewpoint field supplies this initial relationship while $P_{1:T}$ supplies camera intent.
This division allows the same trajectory representation and parameters to serve both viewpoints while leaving their camera--character realization to the learned world prior.

\subsection{Progressive Training Curriculum}
\label{sec:training-curriculum}

Data sources with rich audio-visual semantics and those providing precise, clean control trajectories are fundamentally heterogeneous. To resolve the resulting optimization conflicts, we employ a progressive three-stage curriculum that decouples audio-visual prior acquisition, control alignment, and joint consolidation.

\paragraph{Stage 1: Audio-Visual Continued Pretraining (AV-CPT).}
This stage adapts the base model to the visual styles, motion statistics, and acoustic dynamics (footsteps, ambient audio, and speech) of interactive game and general environments. We optimize the full backbone $\theta$ on the AV-rich data mixture while completely disabling trajectory conditioning. To simultaneously establish text-to-AV generation, image-to-AV generation, and temporal continuation capabilities, we randomly sample the clean media condition $\mathcal{M}$ from an empty set $\varnothing$, a single reference frame $I_{\mathrm{ref}}$, or a synchronized audio-visual prefix $(V_{1:\tau}, A_{1:\tau})$. Unobserved video and audio latent regions are generated jointly, embedding strong acoustic priors and temporal continuity into the backbone.

\paragraph{Stage 2: Action Fine-Tuning (Action-SFT).}
This stage isolates control responsiveness without degrading the acquired audio-visual generation priors. We freeze the entire backbone $\theta$ and train only the lightweight UCPE camera pathway parameters $\phi$ on control-clean data using a visual-only denoising objective. To eliminate information shortcuts, we omit the dynamic-motion narrative and all audio-related fields, retaining only persistent scene, style, viewpoint, and subject descriptors $\mathcal{C}_{\mathrm{static}}$. This forces the trajectory $\mathcal{E}_P$ to serve as the sole source of motion guidance. Furthermore, we heavily bias condition sampling toward reference-frame conditioning ($I_{\mathrm{ref}}$), fixing initial scene content so the network concentrates parameter updates entirely on camera trajectory following.

\paragraph{Stage 3: Joint Fine-Tuning (Joint-FT).}
To consolidate control sensitivity with joint audio-visual synthesis, we unfreeze both the backbone $\theta$ and trajectory pathway $\phi$ for end-to-end co-adaptation on the balanced high-quality mixture. We restore environmental sound and speech fields to the text prompt (while keeping dynamic motion descriptions excluded) and resume joint audio-visual loss optimization. Training executes with a reduced learning rate to harmoniously align multimodal quality with trajectory responsiveness without collapsing the pre-trained priors.

%% file: sections/5_ar.tex
\section{Autoregressive Audio-Visual Generation}
\label{sec:autoregressive-audio-visual-generation}

We progressively adapt a pretrained bidirectional, multi-step audio-visual diffusion model into a streaming autoregressive generator with high throughput and low latency. This transition requires two key transformations: converting bidirectional temporal attention into causal autoregressive computation for online generation of synchronized audio-visual chunks, and compressing iterative denoising into a few-step sampler that remains stable under self-generated histories. To achieve this, we first apply audio-visual teacher forcing~\cite{williams1989learning,chen2024diffusion} to initialize a causal streaming generator from the bidirectional backbone. We then introduce short-horizon Self-Gradient Forcing (SGF)~\cite{huang2026self, zhuang2026sgf} to train on self-generated rollout states while distilling few-step generation, and further extend SGF to long-horizon self-rollouts to improve robustness and stability during sustained streaming inference.

\subsection{Audio-Visual Teacher Forcing}
\label{sec:causal-av-teacher-forcing}

We first use teacher forcing to initialize a causal generator from the
pretrained bidirectional model. Specifically, during training, ground-truth
audio-visual chunks are provided as the historical context, such that the
generation of the current chunk depends only on preceding clean chunks rather
than future observations. This stage provides a stable causal initialization
before exposing the model to self-generated histories in subsequent
self-forcing training.

\paragraph{Causal initialization.}
Let $m\in\{v,a\}$ index the modality, with $v$ denoting video and $a$ denoting
audio. We divide both latent streams into $N$ temporally aligned macro-chunks.
The clean latent of modality $m$ in chunk $i$ is denoted by $x_i^m$, where
$i\in\{1,\ldots,N\}$. An aligned video-audio pair spans the same time interval,
although the two chunks may contain different numbers of tokens. For each pair,
we sample a shared noise level $\sigma_i\in(0,1)$ and construct:
\begin{equation}
  x_{\sigma_i,i}^{m}
  =(1-\sigma_i)x_i^{m}+\sigma_i\epsilon_i^{m},
  \qquad \epsilon_i^{m}\sim\mathcal{N}(0,\mathbf{I}),
  \label{eq:ar-block-noising}
\end{equation}
where $\epsilon_i^m$ is Gaussian noise and $\mathbf{I}$ is the identity covariance
matrix. The shared $\sigma_i$ places the paired audio and video chunks at a
consistent diffusion stage.

\begin{figure}[t]
    \centering
    \includegraphics[width=0.86\linewidth]{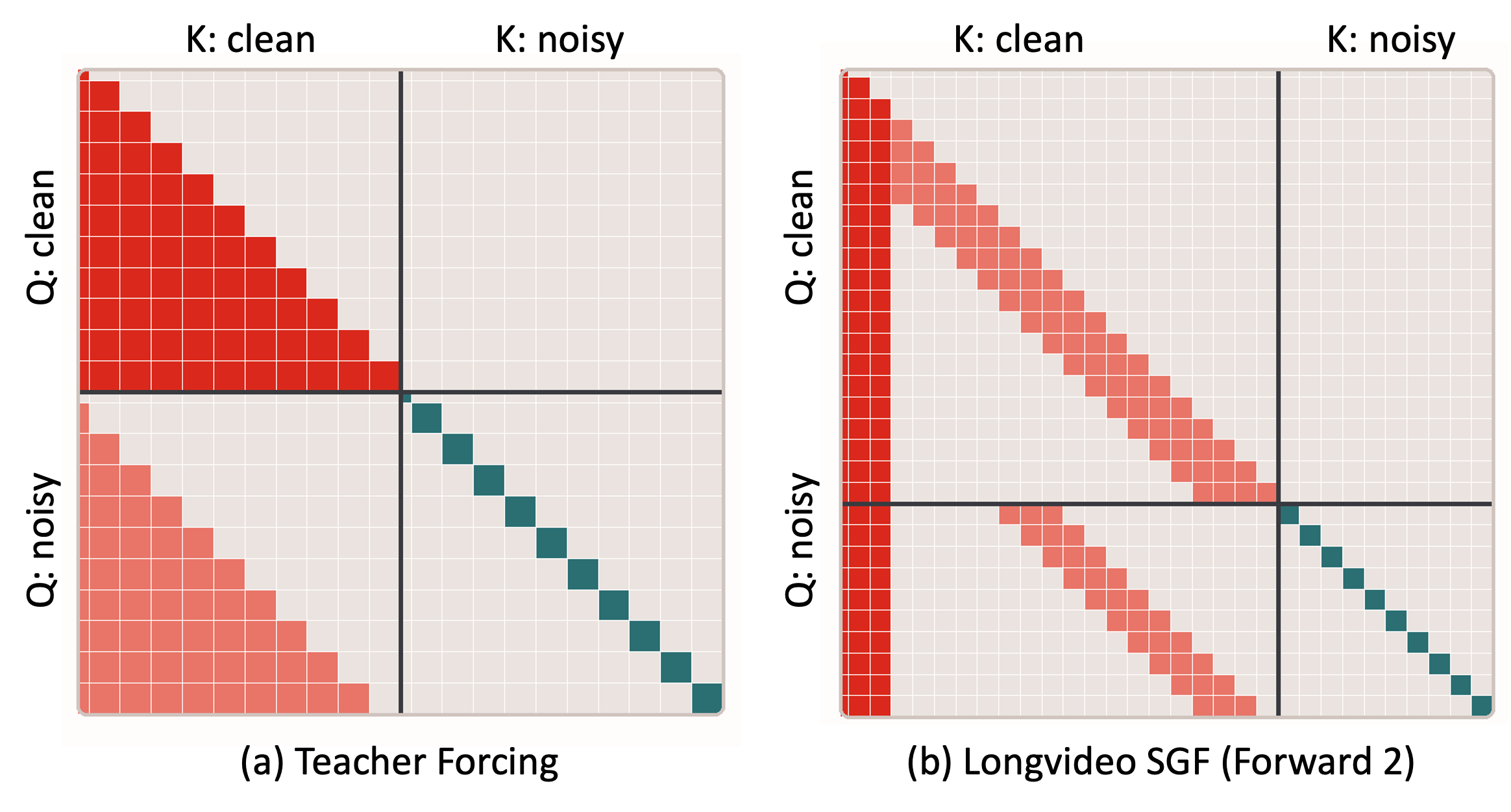}
    \caption{
    Illustration of the causal attention masks used in streaming post-training.
    (a) Audio-visual teacher forcing, where each noisy chunk attends to its
    preceding clean history and the current noisy chunk, while the clean target
    and future chunks are masked out.
    (b) Long-horizon SGF Forward 2, where the causal context is further restricted
    by the sink-plus-FIFO KV cache.
    }
    \label{fig:causal-mask}
\end{figure}

To enable causal generation without sacrificing training parallelism, we
concatenate the clean latents with their noisy counterparts. Let
$\mathcal{B}_{i,\mathrm{clean}}$ and $\mathcal{B}_{i,\mathrm{noisy}}$ denote
the clean and noisy latent blocks of the $i$-th paired audio-visual chunk,
respectively. We impose the following causal attention pattern:
\begin{equation}
\mathcal{S}(\mathcal{B}_{i,\mathrm{clean}})
=
\{\mathcal{B}_{j,\mathrm{clean}}:j\leq i\},
\qquad
\mathcal{S}(\mathcal{B}_{i,\mathrm{noisy}})
=
\{\mathcal{B}_{j,\mathrm{clean}}:j<i\}
\cup
\{\mathcal{B}_{i,\mathrm{noisy}}\},
\label{eq:ar-teacher-forcing-mask}
\end{equation}
where $\mathcal{S}(\cdot)$ denotes the set of latent blocks accessible to the
corresponding attention query. Under this mask, each noisy chunk can attend to
all preceding clean audio-visual chunks and to the noisy tokens within the
current chunk, while the corresponding clean target and all future chunks are
masked out. We apply the same causal constraint to video self-attention, audio
self-attention, and both directions of audio-visual cross-attention. For efficient causal attention computation, we implement this pattern using PyTorch FlexAttention~\citep{dong2025flexattention} with a block-level causal mask. The resulting BlockMask exploits the structured sparsity of chunk-wise causal attention, allowing masked blocks to be skipped without materializing a dense attention mask. 
As illustrated in Fig.~\ref{fig:causal-mask}(a), this causal attention pattern
enables each chunk to use the complete preceding clean history while preventing
information leakage from the current clean target and future chunks.
This converts the original bidirectional temporal attention into chunk-wise
causal attention, while retaining parallel prediction of all chunks during
training.

\paragraph{Training objective.}
To preserve the pretrained generative capability while efficiently adapting
the model to the new causal attention pattern, we perform this stage using
LoRA-based parameter-efficient fine-tuning~\cite{hu2021lora}. The pretrained backbone is kept
fixed, while the LoRA parameters are optimized to learn the causal
audio-visual generation behavior.
 Given the causally masked sequence, the model jointly predicts the flow
velocities of the audio and video latents in a single forward pass. We retain
the standard flow-matching objective of the pretrained diffusion model:
\begin{equation}
  \mathcal{L}_{\mathrm{TF}}
  =
  \sum_{m\in\{v,a\}}\lambda_m
  \mathbb{E}
  \left[
    \sum_{i=1}^{N}
    \left\|
      v_\theta^m(x_{\sigma_i,i}^{m},\sigma_i)
      -
      (\epsilon_i^m-x_i^m)
    \right\|_2^2
  \right],
  \label{eq:ar-teacher-forcing-loss}
\end{equation}
where $v_\theta^m$ denotes the predicted flow velocity for modality $m$,
$\lambda_m$ balances the video and audio losses, and
$\theta$ denotes the trainable LoRA parameters of the generator.
Thus, audio-visual teacher forcing adapts the pretrained bidirectional diffusion
model to causal, chunk-wise autoregressive generation while preserving its
original flow-matching training objective.

\subsection{Short-Horizon Audio-Visual Self-Gradient Forcing}
\label{sec:short-sgf}

Starting from the causal teacher-forcing initialization, we adopt
Self-Gradient Forcing (SGF)~\citep{zhuang2026sgf} to adapt the generator to
its own autoregressive audio-visual histories. SGF follows the deployment-time
self-rollout distribution while restoring gradients through the computation
that encodes generated contexts into future-readable causal memory. Unlike
standard Self Forcing~\citep{huang2026self}, which treats the historical KV
cache as detached rollout state, SGF reconstructs the sampled rollout state
in a differentiable forward pass. Losses on later chunks can therefore
supervise how preceding generated audio-visual contexts are encoded into
causal memory for subsequent generation, without backpropagating through the
full sequential sampling trajectory. We further combine SGF with a DMD
objective~\citep{yin2024onestep} to distill the original multi-step diffusion
model into a few-step autoregressive generator, jointly improving
autoregressive robustness and sampling efficiency.

\paragraph{Forward 1: Autoregressive self-rollout.}
The generator first performs a complete chunk-wise rollout with gradient
computation disabled, using exactly the few-step sampler deployed at inference.
For example, with a four-step sampler, each synchronized audio-visual chunk is
generated through four denoising steps, followed by an additional
clean-context forward at $t=0$ that prefills the causal KV cache for subsequent
chunks. During the rollout, we sample an exit step $t_e$ and record, for each
chunk $i$, the corresponding noisy audio-visual latents
$r_i=(r_i^v,r_i^a)$ together with the final generated clean latents
$\widehat{x}_i=(\widehat{x}_i^v,\widehat{x}_i^a)$. Since Forward 1 is executed
under \texttt{no\_grad}, it faithfully reproduces the inference-time
autoregressive state distribution without retaining the computation graph of
the sequential rollout.

\paragraph{Forward 2: Audio-visual context-gradient reconstruction.}
We then reconstruct the sampled rollout computation in a single differentiable
forward pass using the states recorded in Forward 1. Specifically, the
generated clean latents $\{\widehat{x}_i\}_{i=1}^{N}$ are provided as
stop-gradient context inputs at the clean context timestep $t=0$, while the
recorded noisy audio-visual latents $\{r_i\}_{i=1}^{N}$ retain their sampled
exit timestep $t_e$. The clean-context and noisy target tokens are jointly
arranged according to the causal construction in
Eq.~\eqref{eq:ar-teacher-forcing-mask}: the clean latents of chunk $i$ provide
autoregressive context for subsequent chunks, whereas $r_i$ represents the
prediction target of chunk $i$ at the sampled exit state. This construction
reproduces the context--target relation of the serial rollout while allowing
all recorded chunks to be evaluated in parallel.

Our LTX-2.3-based audio-visual backbone contains five attention pathways that
transmit rollout-dependent information: video self-attention, audio
self-attention, audio-to-video (A2V) attention, video-to-audio (V2A)
attention, and the spatial attention adapter associated with Unified Camera
Positional Encoding (UCPE). As illustrated in
Fig.~\ref{fig:causal-mask}(a), we apply the same chunk-level causal attention
pattern to all five pathways. For a query associated with chunk $i$, the
corresponding keys and values are restricted to the current or preceding
chunks according to the autoregressive rollout geometry. The unimodal masks
prevent video and audio tokens from accessing future content; the A2V and V2A
masks preserve the same temporal causality during bidirectional cross-modal
interaction; and the UCPE mask prevents future camera-conditioned features
from leaking into the current prediction. Text cross-attention remains
unchanged because the text prompt is globally available and contains no
rollout-dependent future information.

Although both the exit-step states and the generated clean latents are
detached from Forward 1, their context representations are recomputed with
gradient tracking in Forward 2. In particular, the $t=0$ clean-context
computation reconstructs the K/V projections and context attention operations
through which generated audio and video chunks are written into causal memory.
Later-chunk losses can therefore propagate through the reconstructed context
representations in all five attention pathways. This supervision improves not
only how each modality writes its own autoregressive history, but also how
audio and video histories are exchanged through A2V and V2A attention and how
camera information is incorporated through UCPE. Applying the matched causal
attention masks allows all chunks to be reconstructed in parallel, yielding
the recovered clean audio-visual predictions
$\{\bar{x}_i=(\bar{x}_i^v,\bar{x}_i^a)\}_{i=1}^{N}$.

\paragraph{DMD optimization.}
We optimize the generator with a joint audio-visual DMD objective evaluated
on the recovered predictions conditioned on self-generated autoregressive
histories. Since these predictions are conditioned on the generator's own
audio-visual history rather than ground-truth context, the objective adapts
the generator to the state distribution encountered during deployment.
Moreover, because the reconstructed predictions are coupled through the
causal A2V and V2A attention pathways, the objective supervises not only the
quality of each modality but also the consistency of their future-readable
representations. It therefore improves few-step sampling efficiency while
mitigating unimodal error accumulation and audio-visual synchronization drift
caused by self-generated histories.

The generator is initialized from the causal model obtained after the
teacher-forcing stage, while both the real-score and fake-score models are
initialized from the pretrained bidirectional diffusion model. The real-score
model is kept frozen to represent the target audio-visual distribution,
whereas the fake-score model is trainable and continuously updated to track
the evolving distribution of the few-step generator.

For a sampled diffusion timestep $s$, we perturb the recovered clean
audio-visual latents as
\begin{equation}
  \bar{x}_{s,i}
  =
  (1-s)\bar{x}_i+s\eta_i,
  \qquad
  \eta_i\sim\mathcal{N}(0,\mathbf{I}),
  \label{eq:sgf-score-noising}
\end{equation}
where $\bar{x}_i=(\bar{x}_i^v,\bar{x}_i^a)$ denotes the jointly recovered
audio-visual prediction. The frozen real-score model
$s_{\mathrm{real}}$ estimates the score of the pretrained data distribution,
while the trainable fake-score model $s_{\mathrm{fake}}$ estimates the score
of the current generator distribution. The generator is optimized with the
DMD objective
\begin{equation}
  \mathcal{L}_{\mathrm{SGF\text{-}DMD}}
  =
  \mathbb{E}_{i,s,\eta_i}
  \left[
    w(s)\,
    \left\langle
      \operatorname{sg}\!\left(
        s_{\mathrm{fake}}(\bar{x}_{s,i},s)
        -
        s_{\mathrm{real}}(\bar{x}_{s,i},s)
      \right),
      \bar{x}_i
    \right\rangle
  \right],
  \label{eq:sgf-dmd-loss}
\end{equation}
where $\operatorname{sg}(\cdot)$ denotes stop-gradient and $w(s)$ is the
diffusion-dependent weighting factor. This objective drives the few-step
generator toward the distribution represented by the pretrained multi-step
model while preserving gradients through the reconstructed causal
audio-visual context.

Following each generator update, we perform multiple fake-score updates. The
generated samples are detached and used to train the fake-score model with the
standard flow-matching objective, allowing it to track the evolving generator
distribution. This short-horizon audio-visual SGF stage establishes robust
few-step generation under deployment-time autoregressive states before
training is extended to substantially longer streaming horizons.

\subsection{Long-Horizon Audio-Visual Self-Gradient Forcing}
\label{sec:long-video-sgf}

Short-horizon audio-visual SGF reconstructs the causal computation within the
original training horizon, but sustained streaming generation encounters
histories containing progressively accumulated prediction errors. We
therefore extend SGF to substantially longer synchronized audio-visual
trajectories. The few-step sampler and DMD formulation remain unchanged; the
main difference is that Forward 1 exposes the generator to long-horizon
self-generated states, while Forward 2 reconstructs the resulting trajectory
under a sparse causal attention pattern. This stage adapts both the denoising
computation and the audio-visual memory-writing process to the distribution
shift encountered during long-running autoregressive generation.

\paragraph{Long-horizon rollout construction.}
We construct a long-horizon trajectory by sequentially rolling out multiple
temporal segments, each corresponding to the training horizon used during
short-horizon SGF. To preserve temporal continuity across segment boundaries,
the terminal video frame of the $k$-th rollout segment is reused as the
initial frame of the $(k+1)$-th segment. Adjacent rollout segments therefore
share a boundary frame instead of restarting from independent initial
conditions. Throughout the trajectory, all subsequent audio-visual chunks are
generated autoregressively from the model's preceding outputs. Consequently,
later segments are conditioned on histories that already contain prediction
errors accumulated during earlier segments.

\paragraph{Bounded multimodal causal context.}
Maintaining a full causal KV cache is impractical for sustained streaming
generation because its size grows with the rollout duration. We instead use a
sink-plus-FIFO policy that preserves a small persistent prefix together with a
bounded window of recent history. This policy is applied consistently to the
video, audio, A2V, V2A, and UCPE attention pathways.

Let $S_m$ denote the number of persistent sink chunks and $R_m$ the number of
recent chunks retained for modality $m$. The historical context directly
available when generating chunk $i$ is
\begin{equation}
  \mathcal{H}_i^m
  =
  \underbrace{
    \bigl(
      \widehat{x}_1^m,\ldots,
      \widehat{x}_{\min(S_m,i-1)}^m
    \bigr)
  }_{\text{attention sink}}
  \mathbin{\Vert}
  \underbrace{
    \bigl(
      \widehat{x}_{\max(S_m+1,i-R_m)}^m,\ldots,
      \widehat{x}_{i-1}^m
    \bigr)
  }_{\text{recent FIFO history}},
  \label{eq:sink-sliding-window}
\end{equation}
where $\mathbin{\Vert}$ denotes ordered concatenation and an indexed range is
empty when its lower bound exceeds its upper bound. The attention sink
preserves a fixed long-term prefix, whereas the FIFO portion continuously
retains the most recent generated chunks. Consequently,
\begin{equation}
  |\mathcal{H}_i^m|
  \leq
  S_m+R_m,
\end{equation}
making the number of historical K/V entries directly accessed at each layer
independent of the total rollout duration. The same sink-plus-FIFO policy and
causal attention geometry are used during training and inference.

As illustrated in Fig.~\ref{fig:causal-mask}(b), Forward 2 reconstructs the
long audio-visual trajectory under the corresponding sink-plus-FIFO masks.
The reconstruction retains the long trajectory as a parallel sparse
computation graph, while each query is allowed to read only the context
specified by its sink-plus-FIFO neighborhood. The clean audio-visual chunks
are recomputed at $t=0$, and the noisy target chunks retain the sampled exit
timestep $t_e$, following the same context--target construction as in
short-horizon SGF.

\paragraph{Layer-wise long-range context gradients.}
Importantly, the sink-plus-FIFO bound constrains only the K/V entries that a
query can read \emph{directly within an individual Transformer layer}. It
does not restrict the end-to-end receptive field or gradient horizon of the
stacked network to the same single-layer window. At an upper Transformer
layer, a recent FIFO chunk visible to the current query already contains
information aggregated from its own sink-plus-FIFO neighborhood in preceding
layers. A loss on the current chunk can therefore first propagate to the
directly visible K/V representations and subsequently continue through their
lower-layer hidden states to the K/V-writing computations of earlier chunks.
Repeating this process across layers creates a multi-hop context-gradient path
whose temporal reach can extend substantially beyond the sink and FIFO entries
directly visible at any single layer.

For example, suppose that chunk $i$ directly attends to a recent chunk $j$ at
layer $\ell$. Although an earlier chunk $h$ may lie outside the direct
sink-plus-FIFO window of chunk $i$, chunk $j$ may have attended to $h$ in a
preceding layer. The gradient from the loss on chunk $i$ can then follow the
path
\begin{equation}
  \mathcal{L}_i
  \longrightarrow
  \mathrm{KV}_{j}^{(\ell)}
  \longrightarrow
  \mathrm{KV}_{h}^{(\ell-1)},
  \label{eq:layerwise-kv-gradient}
\end{equation}
and analogous paths can be composed across additional layers and intermediate
chunks. Hence, $S_m+R_m$ bounds the direct attention fan-in at each layer, but
does not bound the effective multi-layer context-gradient reach.

This layer-wise propagation also operates across modalities. A video loss may
reach earlier audio representations through A2V attention, an audio loss may
reach earlier video representations through V2A attention, and both may
propagate through the camera-conditioned representations introduced by UCPE.
Moreover, the shared boundary between adjacent rollout segments is preserved
during the parallel reconstruction rather than treated as a gradient
boundary. Losses from later segments can consequently supervise more distant
audio-visual context-writing operations through intermediate chunks,
cross-modal attention, and successive Transformer layers.

This transitive supervision remains fundamentally different from full rollout
backpropagation. The generated latents, noise states, sampling decisions, and
serial KV-cache trajectory recorded in Forward 1 remain detached. Gradients
flow only through the parallel Forward-2 reconstruction, including the
clean-context projections, the five causally masked attention pathways, and
the target-side denoising computation. Long-horizon audio-visual SGF therefore
supervises distant memory-writing operations through a bounded-degree sparse
reconstruction graph without retaining the recurrent autograd graph of the
sequential rollout.

\paragraph{Temporal scope of DMD score evaluation.}
During long-horizon training, the autoregressive generator retains its
original RoPE configuration. The sink-plus-FIFO policy only controls which
historical K/V states are retained and visible, without modifying the
generator's positional encoding. Since the real-score and fake-score networks
are initialized from a bidirectional diffusion model pretrained on
fixed-duration sequences, their DMD evaluations during training are restricted
to temporal spans covered by the pretraining horizon. This keeps score-based
supervision within the temporal regime observed by the score models while
leaving the generator's positional representation unchanged.

\paragraph{Long-horizon DMD supervision.}
We apply the DMD objective to every rollout segment instead of supervising
only the final segment of the long trajectory. Let
$\mathcal{L}_{\mathrm{DMD}}^{(k)}$ denote the joint audio-visual DMD loss
computed from the recovered predictions of the $k$-th rollout segment. For a
trajectory containing $K$ segments, the long-horizon objective is accumulated
across all segments:

\begin{equation}
  \mathcal{L}_{\mathrm{Long\text{-}SGF}}
  =
  \sum_{k=1}^{K}
  \mathcal{L}_{\mathrm{DMD}}^{(k)} .
  \label{eq:long-sgf-loss}
\end{equation}

Supervision is therefore distributed throughout the complete self-generated
trajectory. Later-segment losses are evaluated under histories that already
contain errors accumulated in earlier segments and, through the layer-wise
paths described above, can supervise context-writing computations beyond
their directly visible KV windows. Jointly optimizing these segment-wise
losses encourages the few-step audio-visual world model to preserve visual
identity, scene geometry, audio continuity, and cross-modal synchronization
under progressively shifted autoregressive states.

\paragraph{Sparse background temporal regularization.}
We observe that the four-step distilled LTX generator can occasionally
produce transient gray-speckle artifacts in locally stable regions, which
become particularly noticeable during long-horizon autoregressive rollouts.
To distinguish these artifacts from genuine motion, we compare two
consecutive temporal transitions: the historical change from frame $t-2$ to
$t-1$, and the current change from frame $t-1$ to $t$.

Specifically, for each generated latent token $\widehat{z}_{t,p}$ at spatial
location $p$, we first find its closest latent token in a local neighborhood
of frame $t-1$. We then trace this matched token back to its closest
correspondence in frame $t-2$:

\begin{equation}
  q_{t,p}^{*}
  =
  \arg\min_{q\in\mathcal{N}(p)}
  \left\|
    \widehat{z}_{t,p}-\widehat{z}_{t-1,q}
  \right\|_2,
  \qquad
  r_{t,p}^{*}
  =
  \arg\min_{r\in\mathcal{N}(q_{t,p}^{*})}
  \left\|
    \widehat{z}_{t-1,q_{t,p}^{*}}-\widehat{z}_{t-2,r}
  \right\|_2 .
\end{equation}
The corresponding current and historical changes are
\begin{equation}
  d_{t,p}^{\mathrm{cur}}
  =
  \frac{1}{\sqrt{C}}
  \left\|
    \widehat{z}_{t,p}
    -
    \widehat{z}_{t-1,q_{t,p}^{*}}
  \right\|_2,
  \qquad
  d_{t,p}^{\mathrm{hist}}
  =
  \frac{1}{\sqrt{C}}
  \left\|
    \widehat{z}_{t-1,q_{t,p}^{*}}
    -
    \widehat{z}_{t-2,r_{t,p}^{*}}
  \right\|_2,
\end{equation}
where $C$ denotes the latent dimension. If both transitions exhibit
comparable changes, the variation is likely caused by persistent object or
camera motion. In contrast, a current change that is substantially larger
than its historical counterpart is more likely to correspond to a transient
artifact.

To account for textured regions and global frame-level variation, we construct
an adaptive normal-variation scale $s_{t,p}$ by combining the historical
change, the local spatial variation around the matched token, and a stabilizing
frame-wise floor. We select a sparse set $\mathcal{A}$ of locations for which
$d_{t,p}^{\mathrm{cur}}/s_{t,p}$ is unusually large and penalize only the
change exceeding an adaptive tolerance:

\begin{equation}
  \mathcal{L}_{\mathrm{bg\text{-}temp}}
  =
  \frac{1}{|\mathcal{A}|}
  \sum_{(t,p)\in\mathcal{A}}
  \rho\!\left(
    \left[
      d_{t,p}^{\mathrm{cur}}-\alpha s_{t,p}
    \right]_{+}
  \right),
\end{equation}
where $\rho$ is a robust penalty and $\alpha$ controls the tolerated normal
variation. This sparse, motion-adaptive formulation selectively suppresses
transient gray-speckle artifacts while preserving persistent scene, object,
and camera motion. The complete generator objective is
\begin{equation}
  \mathcal{L}_{\mathrm{gen}}
  =
  \mathcal{L}_{\mathrm{Long\text{-}SGF}}
  +
  \lambda_{\mathrm{bg}}
  \mathcal{L}_{\mathrm{bg\text{-}temp}},
\end{equation}
where $\lambda_{\mathrm{bg}}$ controls the strength of the auxiliary
regularization.

\subsection{Autoregressive Inference}
\label{sec:autoregressive-inference}

At inference, the generator produces synchronized audio-visual chunks
autoregressively using the same few-step sampler and sink-plus-FIFO cache
introduced during training. For each new chunk, the retained
modality-specific histories
$\mathcal{H}_i=(\mathcal{H}_i^v,\mathcal{H}_i^a)$ provide the causal context,
while outdated FIFO entries are discarded to keep the KV cache bounded.
Consistent with long-horizon training, the temporal RoPE indices of the sink,
FIFO, and current chunks are rebased into a fixed positional window as
generation proceeds, maintaining a consistent temporal positional layout
throughout streaming inference.

Let $\Theta$ collect the active generator parameters
(including the trajectory pathway $\phi$ for the world-model variant), and let
$\mathcal{D}_i$ denote the variant-specific conditions: text together with
memory and/or an image for the long-video variant, or media context, text, and
trajectory for the world-model variant. The next synchronized chunk is sampled
as:
\begin{equation}
  (V_i,A_i)
  \sim
  p_{\Theta}
  \bigl(\cdot \mid \mathcal{H}_i,\mathcal{D}_i\bigr),
  \label{eq:autoregressive-inference}
\end{equation}
where $V_i$ and $A_i$ denote the generated video and audio chunks,
respectively.

\subsection{Towards Causal Memory}
\label{sec:towards-causal-memory}

Teacher forcing in Eq.~\eqref{eq:ar-teacher-forcing-mask}, long-horizon SGF,
and inference already share one causal rule. A noisy chunk may attend only
to previously generated clean history. The current clean target and all
future chunks remain masked. The sink-plus-FIFO cache in
Eq.~\eqref{eq:sink-sliding-window} is the bounded form of this rule. The
attention sink preserves a fixed early prefix, the FIFO retains the most
recent chunks, and temporal RoPE within each $\mathcal{H}_i^m$ is rebased
as outdated FIFO entries are discarded, so that
$\mathcal{H}_i=(\mathcal{H}_i^v,\mathcal{H}_i^a)$ keeps a consistent
positional layout throughout streaming generation. This cache is sufficient
for local audio-visual continuity under the few-step sampler. It is not
sufficient once a chunk has left the FIFO. Those tokens are no longer
available, even if the sink still holds the opening of the stream. When
the camera returns, or when a model trajectory revisits a previously
observed place, the generator should not treat the scene as new merely
because the corresponding history has been evicted. 

To retain structure beyond the sliding window without changing the causal
mask, the few-step sampler, or the variant-specific conditions
$\mathcal{D}_i$ in Eq.~\eqref{eq:autoregressive-inference}, we introduce a
compact recurrent state $\mathcal{M}_i$ that is written only from the past.
Let $C_i$ denote a short leak-free prefix taken strictly before the start
of chunk $i$, and let $\psi\subset\Theta$ collect the memory parameters
inside the generator, including the trajectory pathway $\phi$ when the
world-model variant is used. The next synchronized chunk and the updated
memory are
\begin{equation}
  (V_i,A_i)
  \sim
  p_{\Theta}
  \bigl(\cdot \mid \mathcal{H}_i,\,\mathcal{M}_{i-1},\,\mathcal{D}_i\bigr),
  \qquad
  \mathcal{M}_i
  =
  \mathcal{F}_{\psi}\bigl(\mathcal{M}_{i-1},\,C_i,\,V_i\bigr).
  \label{eq:causal-memory}
\end{equation}
Thus $\mathcal{H}_i$ continues to bound attention, while $\mathcal{M}_i$
stores what should remain after the FIFO forgets.

\paragraph{Leak-free prefix.}
The construction of $C_i$ follows the same constraint as
Eq.~\eqref{eq:ar-teacher-forcing-mask}. Every index in $C_i$ satisfies
$t<t_i^{\mathrm{start}}$, so the write never observes the current clean
target or any future chunk. The prefix is ordered from oldest to newest
and is concatenated in front of the noisy target, so a left-to-right scan
matches the chunk order of streaming generation. First-chunk windows that
do not contain such a history are dropped during training. Retaining them
forces a first-frame fallback and would allow $\mathcal{F}_{\psi}$ to
condition on a target that is still being denoised, which is the analogue
of letting $\mathcal{B}_{i,\mathrm{noisy}}$ attend to
$\mathcal{B}_{i,\mathrm{clean}}$.

\paragraph{Target-first rebase.}
As the FIFO slides, we already rebase temporal RoPE on $\mathcal{H}_i^m$.
Camera and trajectory encodings that travel with $C_i$ are rebased in the
same way, using the first pose of the current chunk as the origin rather
than the origin of chunk $i-1$. Without this rebase, $\mathcal{M}_{i-1}$
and $C_i$ would remain in the previous chunk's coordinate system, and the
trajectory pathway $\phi$ would observe an artificial jump at every chunk
boundary even when the underlying trajectory is smooth.

\paragraph{Block-wise recurrent write.}
$\mathcal{F}_{\psi}$ is realized as a sparse state-space scan attached to
selected DiT blocks, rather than as a global memory token in text
cross-attention. For each spatial token trajectory the module maintains a
state $s_t$ and updates
\begin{equation}
  s_t
  =
  \sigma(\delta)\odot s_{t-1}
  +
  \bigl(1-\sigma(\delta)\bigr)\odot \widetilde{u}_t,
  \qquad
  y_t
  =
  g_t \odot s_t,
  \label{eq:block-ssm-scan}
\end{equation}
where $\delta$ is a learned per-channel decay, $\widetilde{u}_t$ is a
bounded input projection of the normalized block features, and $g_t$ is a
sigmoid write gate. The scan runs over the concatenated prefix and target,
so the residual at the start of chunk $i$ already contains a summary of
$C_i$. The scan output is added through a residual whose scale is confined
to a small interval. The lower bound prevents the recurrence from
collapsing to a no-op, and the upper bound prevents it from overwriting
the pretrained backbone. After $V_i$ is generated, we retain only the last
few latent-aligned steps as $C_{i+1}$ and take the terminal scan state as
$\mathcal{M}_i$. Long-video variants continue to place text and optional
image conditions in $\mathcal{D}_i$, and world-model variants continue to
condition on $\phi$. Causal memory does not replace those inputs. It is
the component of the history that should remain available after
$\mathcal{H}_i$ slides.

The three components admit natural alternatives. $C_i$ can be replaced by
a suffix, a nearest-first tail, or FoV retrieval. Actions can keep the
previous chunk's coordinates. $\mathcal{F}_{\psi}$ can be omitted, given
an unbounded residual, or trained on first-chunk windows that have no
real prefix. We compared these variants under the same backbone, chunk
length, and SGF schedule. The configuration above, leak-free prefix,
target-first rebase, bounded block-wise write, and history-bearing
training windows, is the one that worked best. It does not change
Eq.~\eqref{eq:autoregressive-inference}. It only specifies what
$\mathcal{H}_i$ contains and how much of that history is written into
$\mathcal{M}_i$ before the sink-plus-FIFO window slides.

%% file: sections/6_agent.tex
\section{Agentic Planning and Control}
\begin{figure}[t!]
\centering
\includegraphics[width=0.9\textwidth]{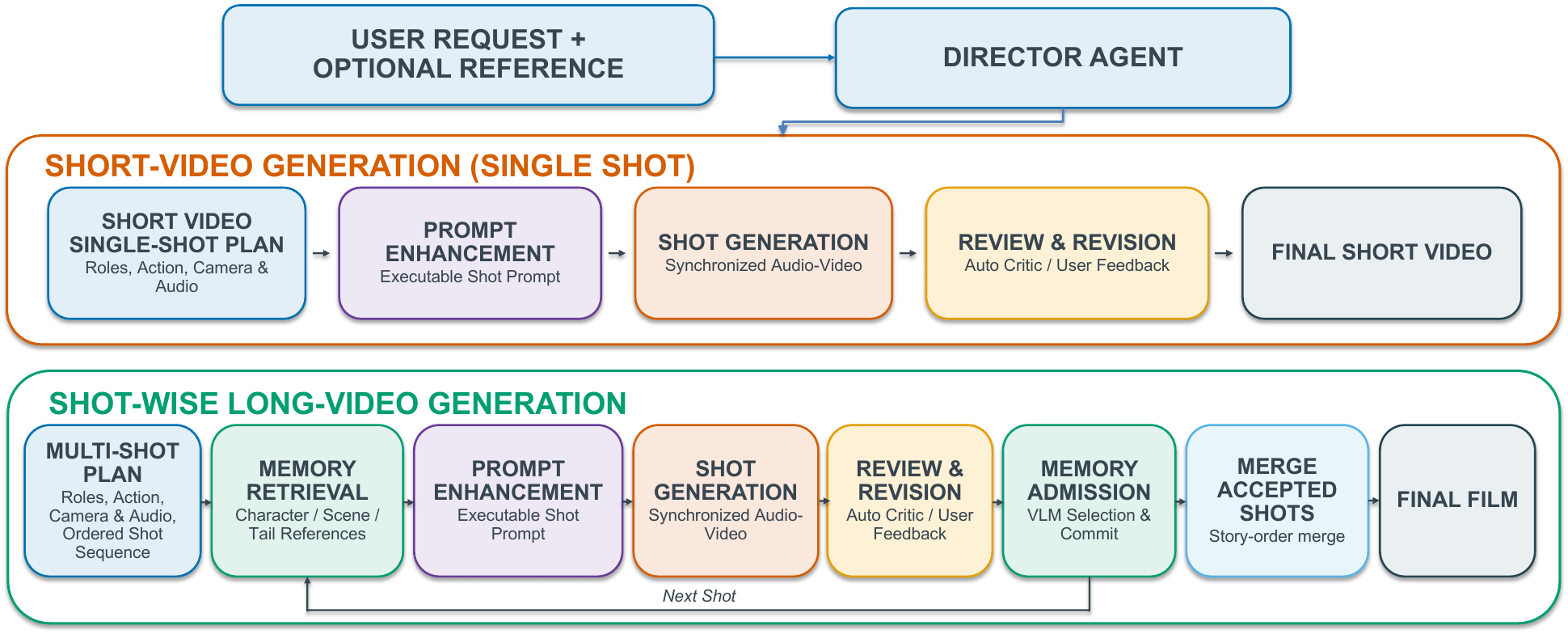}
\caption{Overview of the Director Agent workflow. Given a user request and optional reference, the agent performs shot planning and prompt enhancement (PE), and supports both single-shot short-video generation and iterative long-video generation through memory retrieval, shot-level review and revision, and memory admission.}
\label{fig:agent-workflow}
\end{figure}

An overview of our Director Agent workflow is shown in
Fig.~\ref{fig:agent-workflow}, covering short-video generation and the
shot-wise long-video generation loop with prompt enhancement, memory
retrieval, review, and memory admission.

Long-form video creation begins with an open-ended user intention, whereas a
video generator requires concrete descriptions of what happens in each shot.
A short request rarely specifies a complete narrative, stable character
identities, scene transitions, dialogue, camera progression, or which earlier
visual evidence should be reused.  We therefore introduce a Director Agent that
organizes generation as a persistent, shot-wise process.  The agent develops
the user's intention into a screenplay, applies prompt enhancement (PE),
manages visual--audio memory, and coordinates generation and local revision.
The video model remains responsible for synthesis, while the agent determines
what should be generated and which context should condition it.

A central goal is autonomous completion.  After receiving a brief and any
optional reference media, the agent can understand the request, plan the
story, prepare every shot, generate the synchronized long audio-video sequence,
check the intermediate results, update continuity memory, and assemble the
final film without waiting for a human decision at every shot.  This is the
default autonomous path.  The same path exposes human-in-the-loop checkpoints:
a user can act as a critic, edit the screenplay or current prompt in real time,
reselect a memory, and resume generation from the affected shot.

The workflow is organized around a persistent project workspace.  The
workspace contains the user request, the evolving screenplay, a compact story
profile, the ordered shot plan, generated artifacts, the character and scene
memory banks, and the decisions made during review.  It also records the
relation between neighboring shots and the references supplied to each
generation request.  This state is important for long videos: a later shot is
conditioned on accepted visual evidence and on the user's latest decisions,
rather than on a reconstruction of the conversation or on an unverified plan.
The workspace can therefore be resumed from an intermediate shot, while a
revision changes only the affected shot and its downstream state.

Recent agentic workflows have been explored for long-video generation
\citep{hycreator2026}.
Our workflow separates three responsibilities
that are easy to conflate in a long-video pipeline.  Planning decides what the
story should contain and how it is divided into observable beats.  Generation
turns one prepared beat into synchronized audio and video under the selected
references.  Review and memory admission decide which result is accepted and
which visual evidence is safe to expose to later shots.  Keeping these
decisions explicit makes the workflow inspectable: a failed generation can be
revised without silently changing the screenplay, and an unsuitable memory
proposal can be replaced without altering the accepted shot.

\subsection{Planning and Prompt Enhancement}

\paragraph{Read the brief and develop the screenplay.}
Given a user instruction $u$, the agent first resolves the central event,
characters, locations, temporal progression, and intended ending.  It produces
a complete screenplay before the first generation call, but keeps the
individual shot descriptions editable.  The screenplay establishes the
narrative order, one main dramatic job for each outer shot, and the relation
between adjacent shots.  An outer shot is the unit submitted to the video
generator; any internal camera changes needed to expose one beat remain part
of that shot.  

The planning record contains a compact story profile in addition to prose.  It
gives every recurring subject a stable identity, appearance, wardrobe, and
voice anchor, and records the scene, important props, viewpoint, and relevant
sound context.  Stable subject IDs are reused across all later captions.  This
allows a later mention of a character to refer to an established identity while
still permitting the shot to specify a new pose, action, or camera relationship.
The profile also distinguishes persistent facts from shot-local facts.  For
example, a character's coat and voice can persist, whereas a wet sleeve or a
temporary facial reaction belongs only to the shot in which it is observed.

\paragraph{Prepare the shot plan.}
For every adjacent pair, the planner marks one of three continuity relations:
a scene transition, an ordinary edit within the same scene, or one continuous
action.  A scene transition changes the location or the established spatial
state and starts with fresh scene context.  An ordinary edit keeps the scene
but changes the viewpoint or shot composition.  A continuous action preserves
the terminal state and motion direction of the previous shot.  This relation
is written into the shot plan before generation and is later used consistently
when choosing a representative scene frame, a true tail frame, and the image
references described in the caption.

\paragraph{Construct the production prompt.}
The screenplay and the PE captions are generated as one planning stage:
\begin{equation}
  C=\{c_t\}_{t=1}^{T}
   =\operatorname{PE}\!\left(\mathcal{D}(u)\right).
  \label{eq:agent-plan-pe}
\end{equation}
Here $\mathcal{D}$ produces the narrative plan, $c_t$ is the enhanced caption
for outer shot $t$, and $T$ is the number of planned shots.  We use
$S_t=(V_t,A_t)$ for the synchronized shot, following
Sec.~\ref{sec:unified_memory}.  The equation describes the interface between
planning and generation rather than a new learning objective: the agent writes
the conditions and the video model samples the corresponding shot.

\paragraph{Structured caption fields.}
PE treats a caption as a set of coordinated fields rather than as an
undifferentiated paragraph.  For shot $t$, we write the organization
abstractly as
\begin{equation}
  c_t=\big[r_t^{\mathrm{story}};r_t^{\mathrm{visual}};
  r_t^{\mathrm{camera}};r_t^{\mathrm{audio}};r_t^{\mathrm{temporal}}\big],
  \label{eq:agent-caption-fields}
\end{equation}
where the five terms describe the narrative beat, visible subjects and
environment, camera and composition, synchronized sound, and the ordered
changes within the shot.  The fields are written as natural language in the
final prompt; the notation only makes their roles explicit.

The story field contains the subjects, action, dialogue, and required ending.
The visual field records appearance, clothing, props, scene geometry, lighting,
and style.  The camera field specifies shot scale, viewpoint, composition, and
motivated movement, while the audio field aligns dialogue, Foley, ambience,
and music with the visual event.  The temporal field gives the initial state,
trigger, preparation, continuous action, completion, reaction, and final hold.
The fields preserve our identity anchors, scene relations, and audio--visual
memory roles while keeping the production plan readable at each stage.

\paragraph{Executable detail and grounding.}
PE preserves the hard invariants of the request--subjects, identities, props,
dialogue, language, duration, and outcome--and expands each beat as
initial state $\rightarrow$ trigger $\rightarrow$ action $\rightarrow$
completion $\rightarrow$ reaction $\rightarrow$ final hold.  It adds only
observable evidence that improves execution, such as contact mechanics,
environment response, or material-specific sound; it does not introduce a new
event, subject, or outcome merely to make the caption longer.

\paragraph{Camera, sound, and references.}
The camera field is chosen to expose the evidence: an establishing view gives
geography, a closer view gives expression or manipulation, and a motivated
reframe reveals a consequence.  Dialogue, Foley, ambience, and music follow
the same timeline.  Identity references provide appearance and voice cues,
scene references provide layout and lighting, and an opening tail provides the
starting state for continuous action.  PE names only references supplied to the
current shot, keeping text and media conditions consistent.

\paragraph{Initial and image-aware PE.}
An initial reference image grounds the screenplay in its visible people,
objects, composition, viewpoint, and lighting.  Immediately before an
image-conditioned generation, a second PE pass describes how that state evolves
into the planned beat.  It preserves subject IDs, dialogue, language, and
reference roles while adapting the opening pose and motion.  The same rule is
used for a user-provided first frame and for the tail frame of an accepted shot.
The user may revise the screenplay and shot count before generation; afterward,
the confirmed anchors constrain local prompt edits.

\subsection{Short-Video Generation}

Short-video generation uses the same planning, PE, optional image conditioning,
generation, and user revision procedure as the long-video workflow, but treats
the request as a single-shot screenplay ($T=1$).  The agent still resolves the
event, starting state, ending state, camera observation, and synchronized sound;
the difference is that no later shot needs to inherit the result.  For a fresh
text-to-audio-video request, the enhanced caption is sent directly to the
generator.  The generated audio and video are sampled together so that the
action, speech, Foley, and ambience follow one timeline.

With a user-provided first-frame reference, a multimodal model interprets the
visible pose, viewpoint, environment, lighting, and terminal action state.  It
performs a local image-aware rewrite of the enhanced caption: the opening pose
and state-dependent action are adjusted to follow naturally from the frame,
while the story beat, subject identities, dialogue, language, and reference
roles remain fixed.  The first frame is then passed as the image condition.  A
fresh shot has no image condition; an image-conditioned shot begins from the
provided visual state.  The same interface supports the audio-visual model's
text, image, and memory conditioning without introducing a separate short-video
prompt format.

In autonomous mode, the agent completes the short-video path using its default
review and delivery policy.  In human-in-the-loop mode, the rendered clip is
shown to the user as a critic checkpoint.  The user can provide a concrete
correction--for example, an appearance, action, dialogue, camera, sound, or
continuity change--and the agent rewrites the current caption before
regenerating it.  Since there is no following shot, the accepted short clip
does not update a cross-shot character bank.  This keeps the single-shot path
simple while preserving the same real-time revision semantics used by the
long-video path.

\subsection{Shot-Wise Long-Video Generation}

\paragraph{Autonomous execution.}
The agent executes the confirmed screenplay in timeline order and prepares the
next shot only after the current shot has been resolved.  Auto Mode runs this
loop from the brief to the final merge with default review and memory policies.
The interactive mode exposes the same states to a human critic, who can pause
the current shot, modify its prompt or memory choice, and resume from there.

\paragraph{Memory and shot condition.}
The character memory bank is partitioned by stable character ID,
$B_{t-1}=\{B_{t-1}^{(k)}\}_{k}$, and stores an approved visual observation,
paired audio evidence, and its source shot.  A separate full-frame scene
record preserves same-scene layout; a true tail frame is reserved for
continuous action.  Scene transitions start with fresh scene context.

Before generating shot $t$, the agent retrieves the entries for the identities
in $c_t$.  Let $m_{t-1}^{(k)}$ denote the selected entry for character $k$, and
let $P_t$ denote the same-scene representative; $P_t=\varnothing$ after a
scene transition.  The active memory is
\begin{equation}
  M_{t-1}=\left\{m_{t-1}^{(k)}\mid
    k\in\operatorname{IDs}(c_t)\right\}\cup P_t.
  \label{eq:agent-active-memory}
\end{equation}
Here $\operatorname{IDs}(c_t)$ lists the current subjects.  Character entries
provide identity and speaker cues, while $P_t$ provides scene context.  If the
user requests another existing memory, the agent rebuilds $M_{t-1}$ and
regenerates only the current shot.

\paragraph{Image-aware shot preparation.}
The agent supports fresh generation, first-frame generation, same-scene
reference generation, and continuous generation.  Let $I_t$ be the optional
image condition: a user first frame, the accepted previous tail, or
$\varnothing$ for a fresh shot.  Image-aware PE rewrites only the opening state
needed to connect $c_t$ to $I_t$ and preserves the story beat, subject IDs,
dialogue, and reference roles.  The shot is sampled as
\begin{equation}
  S_t\sim p_\theta\!\left(\cdot\mid
    \widetilde{c}_t,M_{t-1},I_t\right),
  \qquad S_t=(V_t,A_t),
  \label{eq:agent-shot-generation}
\end{equation}
where $\widetilde{c}_t$ is the image-aware caption, $M_{t-1}$ is active memory,
and $I_t$ fixes the starting visual state.

\paragraph{Layered review and real-time intervention.}
After generation, Auto Mode continues with its default policy.  In the
interactive path, the user can accept the clip or provide critic feedback on
appearance, action, dialogue, camera, sound, or continuity.  The agent edits
the current caption or active memory and regenerates only that shot; rejected
artifacts do not enter later context.

\paragraph{Memory admission after shot acceptance.}
After an accepted shot, candidate frames and synchronized audio are collected.
A VLM proposes a subject crop and a scene representative; Auto Mode resolves
the proposal by default, while the user may approve or reselect it.  Character
crops enter the corresponding bank, the scene frame remains continuity-only,
and the true tail is kept for continuous generation.

Let $\widehat{m}_t^{(k)}$ denote the resolved character memory for identity $k$
after the automatic policy or the user's review.  The bank update can then be
written as
\begin{equation}
  B_t^{(k)}=
  \begin{cases}
    \operatorname{Update}\!\left(B_{t-1}^{(k)},\widehat{m}_t^{(k)}\right),
      & \widehat{m}_t^{(k)}\text{ is admitted},\\
    B_{t-1}^{(k)}, & \text{otherwise},
  \end{cases}
  \label{eq:agent-memory-commit}
\end{equation}
where admission is decided automatically or by the user after re-selection.
An unresolved proposal leaves the previous bank unchanged.

\paragraph{Assemble and deliver.}
After memory resolution, the agent prepares the next shot and repeats the loop
until all planned shots are accepted.  It then merges them in screenplay order
with synchronized audio.  Auto Mode delivers the completed film directly;
interactive mode permits a final whole-film review before delivery.

%% file: sections/7_sr.tex
\section{Arbitrary-Step Super-Resolution}
\label{sec:super-resolution}

\begin{figure*}[t]
  \centering
  \includegraphics[width=\textwidth]{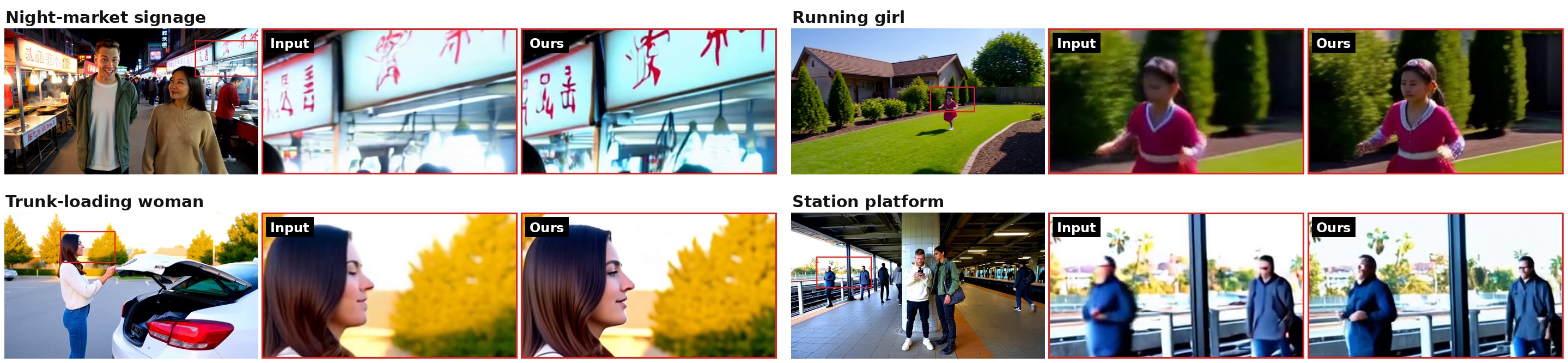}
  \caption{Qualitative effect of the SR stage on generator outputs, concentrated on the two content types that degrade first at the generator's native resolution: small text and small faces. Smeared strokes on distant storefront signage become legible glyphs, and faces only a few dozen pixels across---including under motion---recover defined features rather than the coarse blobs the generator produces. Each cell shows the generator output with the inspected region boxed (left), followed by that region magnified before and after the SR stage.}
  \label{fig:sr-qualitative}
\end{figure*}

The generators of the preceding sections produce synchronized audio-visual content at the 720-class resolutions of the training corpus (Sec.~\ref{sec:hq_data}); a dedicated super-resolution (SR) stage then restores fine texture and suppresses generation artifacts at the delivery resolution. This factorization concentrates the dominant inference cost in the SR stage, which evaluates a large backbone at full output resolution on every frame; with resolution and duration both fixed by the deliverable, the number of network evaluations is the only lever left. Few-step distillation pulls that lever at training time, baking a single step count into the weights~\citep{yin2024onestep,yin2024improved,song2023consistency}, while training-free ODE solvers~\citep{lu2022dpm,zhao2023unipc} stay flexible but degrade sharply in the very-few-step regime that matters most here. We instead distill a flow-matching SR teacher~\citep{lipman2023flow,wan2025wan} with MeanFlow~\citep{geng2025mean} into a model of the \emph{average} velocity over an interval, turning the step count into an inference-time dial: one set of weights serves $N$-step sampling for any $N$, with no retraining and no external solver. Fig.~\ref{fig:sr-qualitative} illustrates the effect of this stage on generator outputs.

\subsection{SR as Conditional Generation}
\label{sec:sr-latent-restoration}

\paragraph{Conditioning by latent concatenation.}
We cast video SR as conditional restoration in the latent space of the Wan2.1 VAE~\citep{wan2025wan}, following the architecture-preserving T2V-to-TV2V super-resolution paradigm of~\citet{luxury2026ultraflash}. The low-resolution (LR) condition occupies the same latent grid as the target, and its latent $z^{\mathrm{LR}}$ is concatenated with the noisy target latent along the channel dimension at the network input, so that the model's task reduces to re-synthesizing the high-frequency content the condition does not contain. To discourage the trivial shortcut of copying the condition, the condition is stochastically corrupted,
\begin{equation}
  y \;=\; (1-\alpha)\,z^{\mathrm{LR}} \;+\; \alpha\,\epsilon_y,
  \qquad
  \alpha \sim \mathcal{U}(0.3,\,0.6),
  \quad
  \epsilon_y \sim \mathcal{N}(0,\mathbf{I}),
  \label{eq:sr-condition}
\end{equation}
with $\alpha$ redrawn on every forward pass; the same perturbation is applied at training and sampling.

\paragraph{AIGC-oriented degradation.}
Because this SR stage is cascaded after a generator rather than applied to camera footage, its condition distribution is dominated by generation artifacts rather than sensor noise. Classical blind-SR pipelines~\citep{wang2021realesrgan,chan2022realbasicvsr} synthesize blur, noise, and compression independently per frame and are therefore mismatched, whereas diffusion-generated video fails through temporally structured modes such as flicker, motion jitter, and color shift. We adopt the AIGC-oriented degradation design of~\citet{luxury2026ultraflash}, prepending temporally coherent degradations whose parameters follow low-frequency trajectories in time, and annealing the overall strength toward a weak final setting matched to the upstream generator. On the target side, an in-batch perceptual-quality gate keeps supervision clean by replacing rejected clips with temporally flipped copies of retained ones.

\paragraph{Text-condition annealing.}
The teacher is adapted from a text-to-video (T2V) backbone into a text-and-video-to-video (TV2V) restorer, and we want the resulting model to super-resolve faithfully \emph{without} relying on a caption at inference: a delivery-time SR stage should not require the upstream prompt to be available, nor should its fidelity depend on how well that prompt is written. Removing the caption from step $0$ is nonetheless the wrong way to obtain this, because the backbone's cross-attention and modulation pathways were pretrained exclusively on caption-conditioned generation; forcing them out of that regime at the same moment the newly added condition pathway is being learned makes two adaptations compete. We therefore adopt the prompt-annealing schedule of~\citet{lu2026omnivr}: the probability of using the real caption decays linearly to zero over the early phase of training, after which a single fixed restoration prompt serves every sample. Annealing separates the two adaptations in time---the text pathway stays in its pretrained regime while gradient signal concentrates on the condition projection, and only then does the text distribution collapse onto the fixed prompt. The resulting model is caption-free at inference while retaining the generative prior that supplies detail the LR input does not contain.

\paragraph{Teacher objective.}
The teacher is trained with the standard conditional flow-matching objective~\citep{lipman2023flow}, consistent with the convention of Eq.~\eqref{eq:ar-block-noising}: for a clean latent $x_0$, noise $\epsilon\sim\mathcal{N}(0,\mathbf{I})$, and noise level $t\in(0,1)$, the noisy latent is $x_t=(1-t)x_0+t\epsilon$ and the network regresses the straight-path velocity,
\begin{equation}
  \mathcal{L}_{\mathrm{SR}}
  =
  \mathbb{E}_{x_0,\,y,\,t,\,\epsilon}
  \left[
    \left\|
      v_\theta\bigl(x_t,\,t,\,y\bigr)
      -
      (\epsilon-x_0)
    \right\|_2^2
  \right],
  \label{eq:sr-teacher-loss}
\end{equation}
where $t$ is drawn from a logit-normal distribution warped by a time-shift transform that concentrates training density at higher noise levels, $y$ is the noise-perturbed condition of Eq.~\eqref{eq:sr-condition}, and the text condition follows the annealing schedule above. The teacher is obtained by approximately $10.4$k optimizer steps of SR fine-tuning through a staged curriculum over degradation strength and resolution, ending on multi-aspect-ratio buckets that match the intended delivery resolutions. At inference, this teacher is a conventional multi-step flow model: producing a single output requires tens of solver steps~\citep{zhao2023unipc}, each a full forward pass at the output resolution, which is precisely the cost that the next subsection removes.

\subsection{Distilling an Arbitrary-Step Sampler}
\label{sec:sr-arbitrary-step}

\paragraph{Average velocity.}
MeanFlow~\citep{geng2025mean} replaces the instantaneous velocity $v(x_t,t)$ with the average velocity over an interval $[r,t]$,
\begin{equation}
  u(x_t,r,t)
  =
  \frac{1}{t-r}
  \int_r^t v(x_\tau,\tau)\,\mathrm{d}\tau,
  \qquad
  0\le r<t\le 1,
  \label{eq:sr-average-velocity}
\end{equation}
so that the exact solution of the probability-flow ODE over $[r,t]$ is the single subtraction $x_r=x_t-(t-r)\,u(x_t,r,t)$. A network $u_\theta(x_t,r,t)$ that models this quantity is an interval integrator by construction: it does not approximate a step of an ODE solver, it \emph{is} the step, for any interval it is queried on.

\paragraph{MeanFlow identity and training objective.}
Differentiating Eq.~\eqref{eq:sr-average-velocity} along the trajectory yields the MeanFlow identity
\begin{equation}
  u(x_t,r,t)
  =
  v(x_t,t)
  -
  (t-r)\,\frac{\mathrm{d}}{\mathrm{d}t}u(x_t,r,t),
  \qquad
  \frac{\mathrm{d}u}{\mathrm{d}t}
  =
  \partial_x u\cdot v + \partial_t u,
  \label{eq:sr-meanflow-identity}
\end{equation}
which converts the intractable interval integral into a local regression target: the known conditional velocity $\epsilon-x_0$ corrected by a Jacobian--vector product (JVP) of the network itself, evaluated under stop-gradient. We train with an explicit-gradient form of this regression,
\begin{equation}
  \mathcal{L}_{\mathrm{MF}}
  =
  \mathbb{E}
  \left[
    \bigl\|
      u_\theta
      -
      \operatorname{sg}\!\bigl(u_\theta + w\,\Delta\bigr)
    \bigr\|_2^2
  \right],
  \qquad
  \Delta
  =
  (\epsilon-x_0)
  -
  u_\theta
  -
  \lambda\,(t-r)\,\widehat{\tfrac{\mathrm{d}u}{\mathrm{d}t}},
  \label{eq:sr-meanflow-loss}
\end{equation}
where $\operatorname{sg}(\cdot)$ denotes stop-gradient and $\widehat{\mathrm{d}u/\mathrm{d}t}$ is a stop-gradient estimate of the total derivative, so that the gradient with respect to $u_\theta$ is $-2w\Delta$ and the update direction is set directly. The per-sample weight $w=1/(\|\Delta\|+c)$ is computed under stop-gradient---a gradient-carrying weight would open a backpropagation path that rewards enlarging the loss to shrink the weight---and down-weights high-noise samples where the JVP estimate is least stable. The tangent norm entering $w$ is rescaled to be invariant to spatial and temporal tensor shape, which keeps a single $c$ valid across all resolution buckets. The factor $\lambda$ linearly ramps from $0$ to $1$ over the first $2{,}000$ steps, interpolating smoothly from pure flow matching to the full MeanFlow objective while the network acquires its interval dependence. The pair $(t,r)$ is drawn as the maximum and minimum of two independent samples from the same shifted logit-normal family used for teacher training (time-shift $2.0$ in the distillation stage), so no new timestep schedule is introduced at distillation time and every sample trains the genuine interval objective ($t>r$).

\paragraph{Jacobian--vector products by finite differences.}
Exact forward-mode JVPs are unavailable in our training stack: fused attention kernels do not implement forward-mode rules, non-reentrant activation checkpointing does not propagate tangents through the recomputed forward, and the sharded data-parallel backward does not support them. We therefore estimate $\mathrm{d}u/\mathrm{d}t$ by a central finite difference \emph{along the ODE trajectory}: perturbing $(x_t,t)$ jointly by $\pm h$ in the direction $(v,1)$ captures both the $\partial_x u\cdot v$ and $\partial_t u$ terms of Eq.~\eqref{eq:sr-meanflow-identity} in a single difference. The two probe evaluations run without gradient tracking, with network outputs kept in full precision and the differencing performed in double precision, since half-precision mantissas are destroyed by the near-cancelling subtraction; both probes share an identical random state and a pinned condition tensor $y$, so that the stochastic conditioning noise of Eq.~\eqref{eq:sr-condition} cannot leak into the difference. One-sided differences are substituted per sample when the interval boundary would be crossed. The step size $h$ was calibrated by sweeping against a higher-precision reference evaluation of the production model---an internal engineering measurement at a single operating point rather than an ablation---and the chosen value sits in the regime where truncation and round-off errors balance; the adaptive weight and warmup in Eq.~\eqref{eq:sr-meanflow-loss} absorb the residual magnitude error, as the objective chiefly consumes the tangent's direction. Measured on our configuration, the estimator costs $2.0\times$ the step time of plain flow matching and no additional activation memory, since peak memory remains set by the single gradient-carrying forward.

\paragraph{Arbitrary-step sampling.}
The distilled model needs no scheduler, no solver, and no per-budget retraining. Given any budget of $K$ network evaluations, we partition the noise axis $1=t_0>t_1>\cdots>t_K=0$ by warping a uniform grid through the same time-shift transform used in training,
\begin{equation}
  t_k
  \;=\;
  \frac{s\,\bigl(1-\tfrac{k}{K}\bigr)}{1+(s-1)\bigl(1-\tfrac{k}{K}\bigr)},
  \qquad k=0,\ldots,K,
  \label{eq:sr-sigma-grid}
\end{equation}
with shift $s=2.0$ and the endpoints fixed to the maximal training noise level and to zero, which concentrates the few available evaluations in the high-noise region where training density was placed, and iterate the exact interval update
\begin{equation}
  x_{t_{k+1}}
  =
  x_{t_k}
  -
  (t_k-t_{k+1})\,
  u_\theta\bigl(x_{t_k},\,t_{k+1},\,t_k\bigr),
  \qquad
  k=0,\ldots,K-1.
  \label{eq:sr-nstep-sampling}
\end{equation}
The sampler consists solely of the grid of Eq.~\eqref{eq:sr-sigma-grid} and the subtraction in Eq.~\eqref{eq:sr-nstep-sampling}; no scheduler object exists. Setting $K=1$ gives one-step generation; larger $K$ trades computation for fidelity along a continuum, and during training we render $K\in\{1,2,4\}$ side by side from the same checkpoint at every validation. This is the deployment property we set out to obtain: the SR cost per output pixel becomes $K$ forward passes with $K$ chosen at request time, so a single model serves latency-critical interactive use at one step and quality-leaning offline rendering at higher step counts, replacing the tens of solver steps required by the teacher. We emphasize what this is not. It is not a fixed-step distilled student in the sense of Secs.~\ref{sec:dmd} and~\ref{sec:short-sgf}, whose DMD objectives~\citep{yin2024onestep,yin2024improved} commit to a specific sampler at training time; and it is not a model to be driven by a multi-step corrector~\citep{lu2022dpm,zhao2023unipc}, which would be both redundant---the subtraction in Eq.~\eqref{eq:sr-nstep-sampling} is already the exact interval integral---and incorrect, since such correctors assume the network predicts the instantaneous velocity $v$, whereas $u_\theta$ predicts the average velocity $u\neq v$.

\subsection{Warm-Starting Without Loss}
\label{sec:sr-warm-start}

\paragraph{The degenerate diagonal.}
MeanFlow contains flow matching as its boundary case: when $r=t$, Eq.~\eqref{eq:sr-average-velocity} reduces to $u(x_t,t,t)=v(x_t,t)$ and the target in Eq.~\eqref{eq:sr-meanflow-loss} degenerates exactly to the flow-matching target of Eq.~\eqref{eq:sr-teacher-loss}. The converged multi-step teacher therefore already solves the MeanFlow objective on the $r=t$ diagonal of the $(r,t)$ plane, and distillation amounts to extending a function the teacher has learned on a line to the triangle above it---not to training a different model that must first re-approach the teacher.

\paragraph{Zero-initialized interval branch.}
To realize this extension without disturbing the teacher, the student architecture augments the backbone with a second conditioning branch that mirrors the timestep pathway: a sinusoidal embedding of the interval length $t-r$, passed through its own MLP and its own modulation projection, whose outputs are \emph{added} to the corresponding timestep-branch activations. Feeding $t-r$ rather than $r$ makes the degenerate case correspond to a fixed zero input, and keeping the projections separate---each branch projected before summation---decouples the new branch's gradients from the pretrained modulation weights. The terminal linear layer of each new pathway is initialized to zero, so at step $0$ the branch contributes exactly nothing and the student's forward pass reproduces the teacher's by construction; a start-of-training self-check verifies that the loss pipeline degenerates to flow matching when $r=t$, confirming the wiring. The warm start is thus lossless: no adapter distillation, no feature matching, and no re-convergence phase are needed.

\paragraph{Curriculum and confound control.}
Distillation is full-parameter and inherits the teacher's entire conditioning stack unchanged---the latent concatenation and the condition noising of Eq.~\eqref{eq:sr-condition}, the degradation pipeline, the paired HR/LR data, and the per-stage resolution buckets---so that any behavioral difference between student and teacher is attributable to the objective rather than to a data-distribution shift. Training proceeds in two stages: a first stage on moderate-resolution buckets initialized from the converged teacher, followed by a continuation on the teacher's own high-resolution multi-aspect buckets with the optimizer state carried over; the shape-invariant tangent normalization of Sec.~\ref{sec:sr-arbitrary-step} is what allows this resolution change without retuning the loss weighting. The optimizer follows the teacher's settings except for a raised second-moment coefficient ($\beta_2$ from $0.95$ to $0.99$), accommodating the higher gradient variance of the finite-difference target, and the high-resolution stage adds a cross-rank finiteness gate that skips optimizer steps on non-finite gradients before clipping, preserving the Adam state through transient instabilities and closing a failure mode in which a NaN gradient silently bypasses standard norm clipping because the comparison \texttt{nan < 1.0} evaluates to false. Combined with the tangent warmup, the overall trajectory is a continuous deformation of the teacher: the student begins as the teacher, is annealed onto the MeanFlow objective, and ends as a single set of weights for which Eq.~\eqref{eq:sr-nstep-sampling} holds at every step budget.

%% file: sections/8_eval.tex
\section{Results}

\begin{table*}[t]
\centering
\caption{Quantitative comparison of baseline methods. ViCLIP and Self-CIDS assess cross-shot visual and identity consistency, Voice assesses speaker consistency, and Recall measures speech-content coverage. All metrics are higher-is-better. Boldface and underlining indicate the best and second-best results, respectively.}
\label{tab:baseline_comparison}
\small 
\resizebox{\linewidth}{!}{
\begin{tabular}{l|ccc|cc|c|c}
\toprule
 & \multicolumn{3}{c}{Cross-Shot Consistency} & \multicolumn{2}{c}{Video Quality} & \multicolumn{1}{c}{Text Consistency} & \multicolumn{1}{c}{Speech} \\
\cmidrule(lr){2-4} \cmidrule(lr){5-6} \cmidrule(lr){7-7} \cmidrule(lr){8-8}
Method & ViCLIP $\uparrow$ & Self-CIDS $\uparrow$ & Voice $\uparrow$ & Aesthetic $\uparrow$ & Imaging $\uparrow$ & CLIP $\uparrow$ & Recall $\uparrow$ \\
\midrule
JavisDiT++~\citep{liujavisdit++} & 0.6955 & 0.5621 & 0.7933 & 0.5019 & 0.6084 & 0.2522 & 0.0951 \\
LTX-2.3~\citep{hacohen2026ltx2} & 0.7047 & 0.6135 & 0.6847 & 0.5436 & 0.4751 & {0.2559} & 0.9304 \\
LTX-2.5~\citep{hacohen2026ltx2} & 0.7775 & 0.6640 & 0.7603 & \textbf{0.6019} & 0.5804 & \underline{0.2683} & \underline{0.9545} \\
ShotStream+MMAudio~\citep{luo2026shotstream} & {0.7988} & 0.7308 & {0.7945} & 0.5255 & 0.6044 & 0.1998 & 0.0297 \\
StoryMem+MMAudio~\citep{zhang2025storymem} & 0.7897 & {0.7491} & 0.7643 & 0.5265 & {0.6320} & {0.2368} & 0.0301 \\
HappyOyster (Directing) & 0.7535 & 0.6940 & 0.7705 & {0.5606} & {0.6701} & {0.2544} & 0.1085 \\
JoyAI-Echo-1.0~\citep{li2026joyai} & \underline{0.8026} & \underline{0.7793} & \underline{0.8129} & 0.5679 & \underline{0.7058} & 0.2658 & 0.9489 \\
\midrule
JoyAI-Echo-1.5 & \textbf{0.8264} & \textbf{0.7937} & \textbf{0.8524} & \underline{0.5705} & \textbf{0.7467} & \textbf{0.2868} & \textbf{0.9674} \\
\bottomrule
\end{tabular}
}
\vspace{1mm}
\end{table*}

We evaluate JoyAI-Echo-1.5 from two complementary perspectives:
\emph{long-horizon audio-visual generation} and
\emph{action-conditioned world modeling}.
For world modeling, we report results on two public benchmarks,
WBench and SANA-WM-Bench, together with a large-scale human preference study.

\subsection{Long-Horizon Audio-Visual Generation}

\noindent\textbf{Experimental Setup.}
We evaluate long-horizon audio-visual generation on a curated benchmark of
100 stories and 3,000 shots, following the protocol of
JoyAI-Echo-1.0~\citep{li2026joyai}. We consider four complementary aspects:
\begin{itemize}
\item \textbf{Cross-shot consistency.}
We report pairwise ViCLIP similarity~\citep{wanginternvid} for
visual-semantic consistency, Self-CIDS for character identity consistency
using GroundingDINO~\citep{ShilongLiu2023GroundingDM},
Torchreid~\citep{torchreid}, and FaceNet~\citep{schroff2015facenet},
and 3DSpeaker similarity~\citep{chen20243d} for speaker consistency.
\item \textbf{Video quality.}
Following VBench-Long~\citep{huang2024vbench}, we concatenate all
generated shots into a long video, split it into 2-second clips, and
evaluate aesthetic and imaging quality.
\item \textbf{Text consistency.}
We measure alignment between the generated visual content and the input
story script using CLIP score~\citep{radford2021learning}.
\item \textbf{Speech-content fidelity.}
We transcribe generated speech with Whisper~\citep{radford2022whisper}
and compute word-level recall against the reference dialogue. We use
recall instead of the accuracy metric in JoyAI-Echo-1.0 and re-evaluate
all methods under the same protocol.
\end{itemize}

Unless otherwise specified, all JoyAI-Echo-1.5 results are generated with the
Director Agent enabled for shot-level planning and memory selection.

\noindent\textbf{Baselines.}
We compare JoyAI-Echo-1.5 with three classes of audio-visual generation
systems:
\begin{itemize}
\item \textbf{Joint audio-visual generation:}
JavisDiT++~\citep{liujavisdit++}, LTX-2.3, and
LTX-2.5~\citep{hacohen2026ltx2}. Each method generates approximately
10-second shots from the corresponding structured prompts, which are then
concatenated for long-form evaluation. For LTX-2.3 and LTX-2.5, we enable
prompt enhancement and condition each shot after the first on the final
frame of the preceding shot.
\item \textbf{Cascaded generation:}
ShotStream~\citep{luo2026shotstream}+MMAudio~\citep{cheng2025mmaudio}
and StoryMem~\citep{zhang2025storymem}+MMAudio. The video model first
generates a silent 10-second shot, after which MMAudio synthesizes audio
from the corresponding shot-level audio description. The resulting shots
are concatenated using the same story scripts and shot durations.
\item \textbf{Native long-form generation:}
HappyOyster in Directing mode,\footnote{Details available at
\url{https://www.happyoyster.cn/docs}} which directly generates
continuous audio-visual sequences from narrative scripts. 
\end{itemize}

\begin{figure}[t!]
\centering
\includegraphics[width=1.0\textwidth]{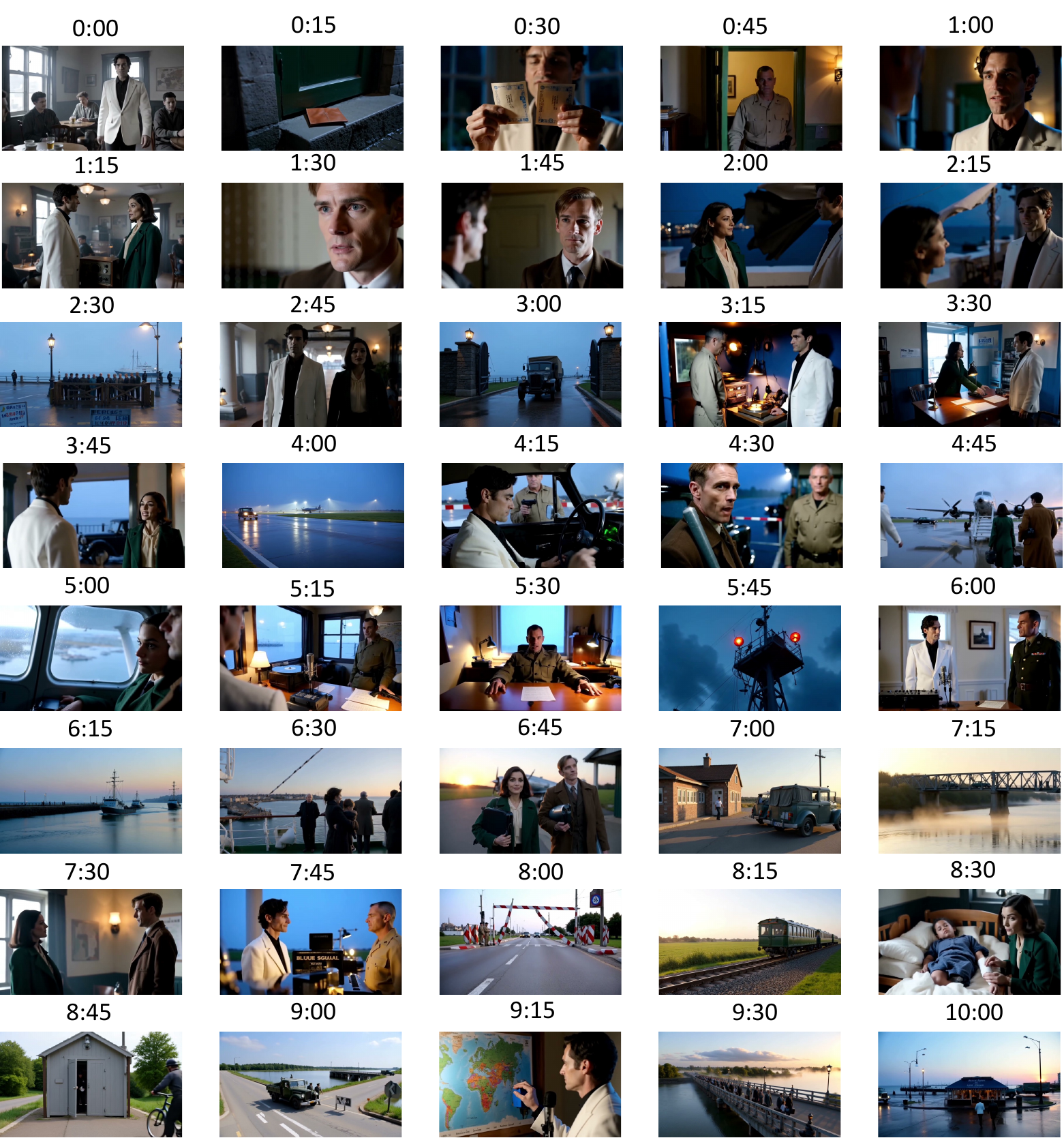}
\caption{Qualitative visualization of a 10-minute long audio-visual sequence generated under the Director Agent. The 10-minute example follows a wartime evacuation story set in a neutral port city. A radio operator helps two civilians escape as military control tightens and the city gradually descends into chaos. Their escape becomes increasingly entangled with checkpoints, shifting alliances, and the growing danger faced by civilians left behind. The radio operator eventually turns his broadcast network into a broader evacuation effort, connecting scattered groups and helping them find a route out of the city.}
\label{fig:compare-10min}
\end{figure}

\noindent\textbf{Quantitative Comparison.}
Tab.~\ref{tab:baseline_comparison} compares JoyAI-Echo-1.5 with native
audio-visual generators, cascaded long-video systems, and JoyAI-Echo-1.0.
JoyAI-Echo-1.5 achieves the best performance on six of the seven metrics and
ranks second on aesthetic quality. In particular, it obtains the highest
ViCLIP (0.8264), Self-CIDS (0.7937), and Voice (0.8524) scores, improving over
JoyAI-Echo-1.0 by 0.0238, 0.0144, and 0.0395, respectively. These results
demonstrate substantially improved cross-shot visual and speaker consistency.
Meanwhile, JoyAI-Echo-1.5 achieves the best Imaging (0.7467), CLIP (0.2868),
and speech Recall (0.9674), while remaining competitive in aesthetic quality.
This indicates that the gains in long-range consistency are achieved without
sacrificing local generation quality, text alignment, or speech fidelity.

\noindent\textbf{User Study.}
We further conduct a human evaluation comparing the long-video variant of
JoyAI-Echo-1.5 with HappyOyster in Directing mode through randomized and
anonymized side-by-side comparisons. Following the Good--Similar--Bad (GSB)
protocol, annotators evaluate story-instruction following,
dialogue/background-audio following, memory/identity consistency,
audio-visual synchronization, and audio quality. For each dimension, they
determine whether JoyAI-Echo-1.5 is better, the two results are similar, or
HappyOyster is better.

As shown in Tab.~\ref{tab:gsb_happyoyster}, JoyAI-Echo-1.5 receives a higher
preference rate than HappyOyster across all five dimensions. The largest
advantage is observed in audio-visual synchronization, where JoyAI-Echo-1.5
is preferred in 48.4\% of the comparisons, compared with 20.3\% for
HappyOyster. JoyAI-Echo-1.5 also demonstrates clear advantages in
story-instruction following (38.6\% vs.\ 17.7\%) and
dialogue/background-audio following (39.2\% vs.\ 23.5\%).

For memory/identity consistency, JoyAI-Echo-1.5 is preferred in 24.8\% of
the comparisons, compared with 15.7\% for HappyOyster, while 59.5\% are
judged similar. This indicates that both systems maintain identity adequately
in many cases, but JoyAI-Echo-1.5 is more frequently preferred when
perceptible differences occur. JoyAI-Echo-1.5 also achieves a higher
preference rate in audio quality (44.4\% vs.\ 33.3\%). Overall, the user
study demonstrates that the improvements measured by the automatic metrics
remain perceptible in complete long-video generation, particularly in
instruction following, audio-visual synchronization, and long-range
consistency.

\begin{figure}[t!]
\centering
\includegraphics[width=1.0\textwidth]{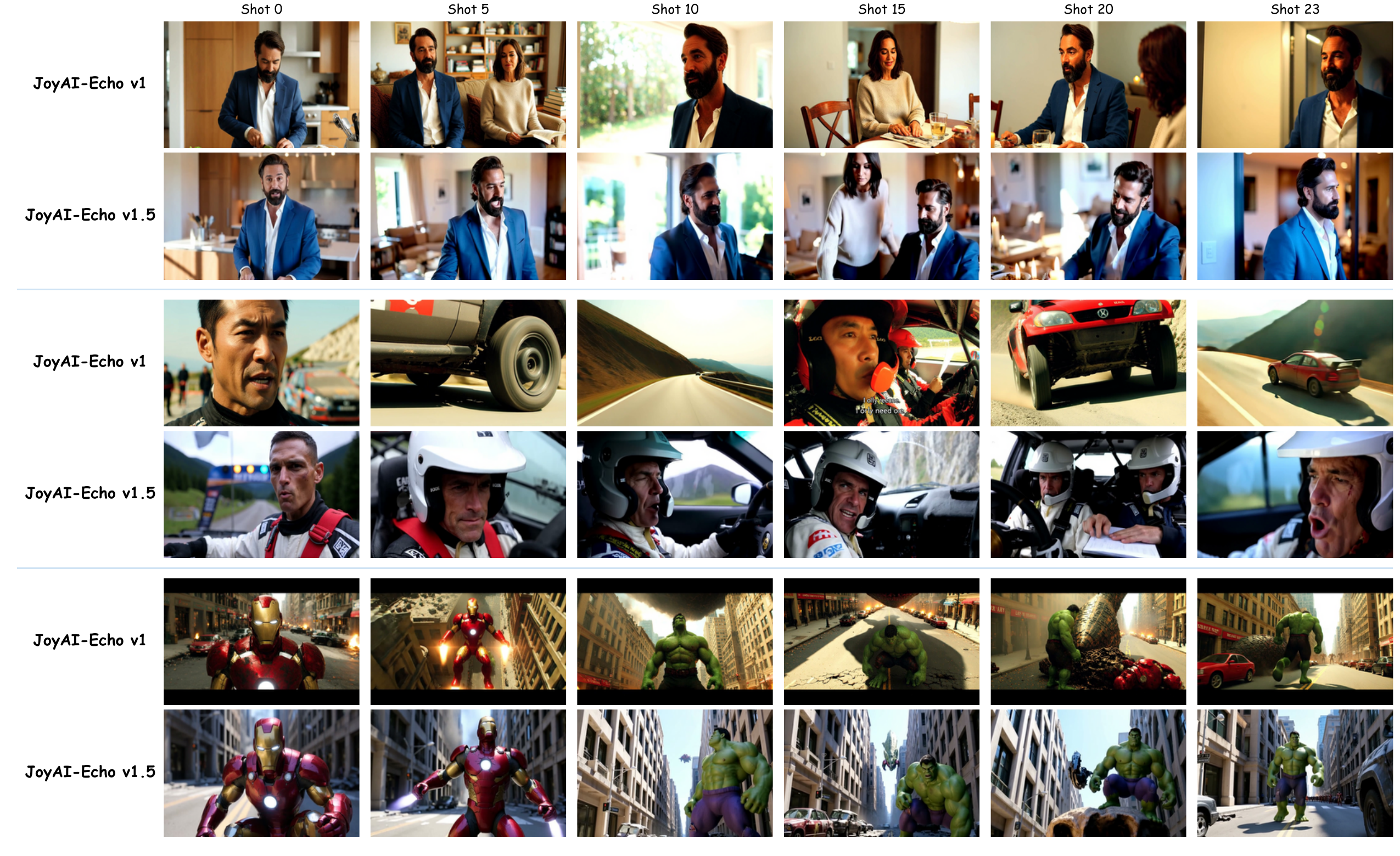}
\caption{Qualitative comparison between JoyAI-Echo-1.5 and JoyAI-Echo-1.0 under
the same long-form story. Frames sampled from temporally separated shots
demonstrate the improved character consistency, scene stability, and visual
rendering of JoyAI-Echo-1.5.}
\label{fig:compare-longvideo}
\end{figure}

\begin{table}[t]
\centering
\small
\caption{
Human evaluation based on Good--Similar--Bad (GSB) pairwise comparisons
between JoyAI-Echo-1.5 (long-video variant) and HappyOyster for long-horizon audio-visual
generation. Numbers denote the percentage of user preferences and may not sum
to exactly 100\% because of rounding.
}
\label{tab:gsb_happyoyster}
\begin{tabular}{lccc}
\toprule
Aspect
& JoyAI-Echo-1.5
& Similar
& HappyOyster \\
\midrule
Story-instruction following
& \textbf{38.6\%} & 43.8\% & 17.7\% \\
Dialogue/background-audio following
& \textbf{39.2\%} & 37.3\% & 23.5\% \\
Memory/identity consistency
& \textbf{24.8\%} & 59.5\% & 15.7\% \\
Audio-visual synchronization
& \textbf{48.4\%} & 31.4\% & 20.3\% \\
Audio quality
& \textbf{44.4\%} & 22.2\% & 33.3\% \\
\bottomrule
\end{tabular}
\end{table}

\noindent\textbf{Qualitative Comparison.}
Fig.~\ref{fig:compare-longvideo} compares JoyAI-Echo-1.5 with
JoyAI-Echo-1.0 using the same long-form story and shot-level instructions.
Compared with JoyAI-Echo-1.0, JoyAI-Echo-1.5 more reliably preserves the
facial structure, hairstyle, clothing cues, and speaker identity of recurring
characters across temporally distant shots. It also exhibits more stable
exposure and color balance, with less accumulation of oversaturation,
contrast artifacts, and scene drift over extended generation. These
improvements are particularly visible after multiple scene changes, camera
scale variations, and the introduction of new characters.
In addition to the benchmark evaluation, we present a separate 10-minute
full-system case study in Fig.~\ref{fig:compare-10min}. Despite the extended temporal horizon and frequent shot transitions, the generated sequence maintains recognizable character identities, coherent scene evolution, and stable visual quality across distant timestamps. This example demonstrates the system's ability to sustain consistent long-form generation far beyond the duration of individual shots.

These examples illustrate the complementary roles of the model and the
long-video generation pipeline. At the generator level, unified audio-visual
Memory conditioning provides persistent identity and speaker cues while still
allowing each target shot to follow its own textual description. The
high-quality training corpus improves local rendering quality and broadens the
model's spatial and linguistic coverage, while transition-aware multi-shot
captions strengthen its ability to represent changes in subject, action,
camera view, and scene context around shot boundaries. At the system level,
the Director Agent selects grounded Memory observations and revises the
shot-level plan before subsequent generation, reducing the propagation of
incorrect characters, actions, or scene attributes.
Compared with JoyAI-Echo-1.0, JoyAI-Echo-1.5 therefore exhibits improvements
at both local and long-range scales. Individual shots show clearer structure
and more balanced rendering, while the full sequence exhibits less character
drift, scene drift, and color accumulation over time. The qualitative results
are consistent with the improvements in imaging quality, CLIP alignment,
cross-shot visual consistency, and speaker consistency reported in
Tab.~\ref{tab:baseline_comparison}. A more complete long-horizon visualization
with densely sampled timestamps is provided in the supplementary material.

\input{sections/08_sec02_eval_causal}

%% file: sections/08_sec02_eval_causal.tex

\subsection{Action-Conditioned World Modeling}

We evaluate two world-model variants of JoyAI-Echo-1.5:
the original undistilled bidirectional multi-step model, denoted as
\textbf{JoyAI-Echo-1.5}, and its distilled causal four-step variant, denoted as
\textbf{JoyAI-Echo-1.5-Causal}.
We benchmark these models on two public benchmarks,
WBench~\citep{ying2026wbench} and
SANA-WM-Bench~\citep{zhu2026sanawm}, together with a subjective evaluation on
our self-built World-Model Benchmark (WMB).
The former evaluates broad interactive-world behavior, while the latter
focuses on camera-trajectory following and long-horizon visual persistence.

\input{figures/compare_wbench}

\input{figures/fig_causal_wbench_comparisons}
\input{figures/fig_causal_result_60s}

\subsubsection{WBench}

\paragraph{Evaluation settings and metrics.}
We evaluate both JoyAI-Echo-1.5 and JoyAI-Echo-1.5-Causal on the official
158-case Navigation split of WBench~\citep{ying2026wbench}, which covers
first- and third-person interactive navigation.
Following the official multi-turn evaluation protocol, we report five
dimensions---Quality, Setting, Interaction, Consistency, and Physical---together
with their overall Average score. Quality aggregates perceptual and temporal
video-quality metrics; Setting evaluates scene and subject adherence;
Interaction measures compliance with the requested interaction, including
navigation trajectory following; Consistency aggregates spatial, geometric,
photometric, perspective, subject, and segment-level consistency; and Physical
measures visual plausibility and causal fidelity.

\paragraph{Quantitative Results.}

\input{tables/tab_wbench_navi}

As shown in Tab.~\ref{tab:wbench-navigation}, both variants of our model rank
at the top of the WBench Navigation leaderboard. The undistilled bidirectional
multi-step JoyAI-Echo-1.5 achieves the highest overall Average score of 81.7,
while the distilled four-step JoyAI-Echo-1.5-Causal ranks second with 81.0.
This small performance gap indicates that causal few-step distillation largely
preserves the interactive world-modeling capability of the original multi-step
model.

JoyAI-Echo-1.5 achieves a Consistency score of 89.8, the second highest among
all evaluated models, demonstrating strong world-state preservation across
multi-turn interactions. Notably, JoyAI-Echo-1.5-Causal attains the highest
Interaction score of 87.9, slightly surpassing the undistilled model at 87.2.
Despite the transition from bidirectional multi-step generation to four-step
causal autoregressive inference, the distilled model therefore retains strong
control responsiveness while remaining competitive in overall generation
quality.

\paragraph{Qualitative Results.}

Fig.~\ref{fig:wbench-comparison-3} first compares the undistilled
bidirectional multi-step JoyAI-Echo-1.5 with representative world models on
WBench Navigation cases. Given the same initial observation and navigation
condition, competing methods often exhibit abrupt viewpoint changes, unintended
zooming, subject-scale variation, layout drift, or loss of scene identity.
In contrast, JoyAI-Echo-1.5 produces more coherent world evolution across
diverse visual domains, with stronger navigation adherence and better
preservation of scene structure and appearance. These observations are
consistent with its strong Interaction and Consistency scores in WBench.

We next compare causal few-step world models in
Fig.~\ref{fig:wbench-causal-comparison}, including DreamX-World,
LingBot2-Fast, SANA-WM Streaming with and without Causal Refiner, Evoke,
and our JoyAI-Echo-1.5-Causal. We evaluate three representative multi-turn
interaction cases:
\texttt{A} $\rightarrow$ \texttt{A} $\rightarrow$
\texttt{A} $\rightarrow$ \texttt{A},
\texttt{Left} $\rightarrow$ \texttt{Left} $\rightarrow$
\texttt{Right} $\rightarrow$ \texttt{Right}, and
\texttt{Left} $\rightarrow$ \texttt{W} $\rightarrow$
\texttt{Right}, respectively. Here, \texttt{Left}, \texttt{Right},
\texttt{Up}, and \texttt{Down} denote camera rotations, whereas
\texttt{W}, \texttt{S}, \texttt{A}, and \texttt{D} denote translational
movements. Among the representative baselines, LingBot2-Fast and Evoke remain
relatively competitive across multiple interactions, yet still exhibit
noticeable action deviation together with gradual degradation in subject
identity and visual style as the rollout progresses. Other baselines show more
pronounced viewpoint drift, structural changes, or accumulated appearance
inconsistency. In contrast, JoyAI-Echo-1.5-Causal follows the prescribed
controls more accurately while better preserving scene layout, subject
identity, and visual style across successive turns, demonstrating stronger
multi-turn extrapolation and improved robustness to accumulated autoregressive
errors.

We further examine the long-horizon extrapolation capability of the distilled
causal model in Fig.~\ref{fig:causal-wbench}. Over approximately 60 seconds
of continuous causal streaming, JoyAI-Echo-1.5-Causal remains responsive to
navigation controls while largely preserving scene identity, subject appearance,
visual style, and major spatial structure across successive generated segments.
The sustained coherence over this substantially extended horizon further
indicates that the causal generator is robust to autoregressive error
accumulation well beyond the training horizon.


\subsubsection{SANA-WM-Bench}

\paragraph{Evaluation settings and metrics.}
We evaluate SANA-WM-Bench under three complementary settings:
(1) a short-horizon evaluation on the first 241 frames of each trajectory,
(2) the undistilled multi-step model under the official 961-frame
(60-second at 16 FPS) long-horizon protocol, and
(3) the distilled causal model at 961 frames.
The first setting provides a short-horizon reference for trajectory following
and visual quality, the second follows the official long-horizon protocol to
characterize the 60-second generation capability of the undistilled model, and
the third evaluates whether the distilled few-step causal generator can retain
strong long-horizon performance after distillation.
For trajectory fidelity and visual quality, we follow the benchmark protocol
and report rotation error ($R$, in degrees), relative translation error ($T$),
camera-motion consistency error (CMC), and VBench Overall on a 0--100 scale.
For the 961-frame evaluations, we additionally measure viewpoint revisit
consistency using PSNR, SSIM, and LPIPS between generated frame pairs that
revisit approximately the same camera pose, thereby evaluating long-term scene
persistence without reference to ground-truth frames. We further report the
temporal quality degradation
$\Delta\mathrm{IQ}=q_1-q_6$, where $q_1$ and $q_6$ denote the VBench Imaging
Quality scores of the first and last 10-second windows, respectively.

\paragraph{Quantitative Results.}

\input{tables/tab_sana_wm_241}

\input{tables/tab_sana_wm_961}
\input{tables/tab_sana_wm_961_causal_few_steps}
\input{figures/fig_sana_iq_summary}

As shown in Tab.~\ref{tab:sana-wm-bench-241}, in the 241-frame
short-horizon setting, JoyAI-Echo-1.5 achieves the highest VBench Overall on
both trajectory splits, reaching 83.91 on Simple and 83.96 on Hard. It also
obtains the lowest translation and CMC errors on the Simple split, while
remaining close to SANA-WM in rotation accuracy. On the Hard split, SANA-WM
achieves lower trajectory errors, whereas JoyAI-Echo-1.5 maintains the highest
overall visual quality.

As shown in Tab.~\ref{tab:sana-wm-bench-961}, under the official 961-frame
long-horizon setting, the undistilled multi-step JoyAI-Echo-1.5 remains highly
competitive in both trajectory following and scene persistence. On the Simple
split, it achieves a VBench Overall of 81.36, the lowest translation and CMC
errors, and the highest revisit PSNR of 15.10,dB. Although trajectory errors
become more pronounced over the long rollout, the model maintains strong visual
quality and revisit consistency over the full 60-second trajectory. On the
Simple split, its rotation error reaches $3.22^\circ$ at 961 frames, while on
the Hard split it increases to $12.05^\circ$, indicating that accumulated pose
drift remains a key challenge under difficult long-horizon trajectories.

Tab.~\ref{tab:sana-wm-bench-causal-961} further compares the distilled causal
model with recent state-of-the-art causal distillation methods under the same
961-frame horizon. JoyAI-Echo-1.5-Causal achieves the highest VBench scores on
both Simple and Hard splits, with 80.13 and 81.06, respectively, together with
the best trajectory accuracy and strongest revisit consistency among the
evaluated causal models. As further illustrated in
Fig.~\ref{fig:iq-comparison}, it achieves substantially higher first-window,
mean, and minimum IQ than Evoke and SANA-WM with Causal
Refiner, while maintaining competitive final-window IQ. Although these
baselines exhibit smaller $\Delta$IQ, their lower initial and overall IQ suggest
that a small drop may also result from consistently low imaging quality rather
than stronger preservation of a high-quality initial state. Compared with the
undistilled 961-frame model, causal few-step distillation introduces a moderate
degradation in trajectory accuracy and revisit consistency while largely
preserving long-horizon visual quality, demonstrating robust 60-second
autoregressive rollout capability.

\paragraph{Qualitative Results.}

\input{figures/fig_sana_causal_results_comparisons}

Fig.~\ref{fig:sana-causal-qualitative} compares approximately 60-second
causal rollouts generated from the same initial observation under matched
camera trajectories. As the rollout horizon increases, competing methods
exhibit different forms of cumulative autoregressive drift.
LingBot-World-v2 progressively alters the scene composition and viewpoint,
while SANA-WM without the Causal Refiner can undergo substantial appearance
and style shifts. The Causal Refiner improves local temporal stability but
does not fully prevent long-term structural and appearance drift. Evoke
generally maintains coherent short-term appearance, but deviations in
viewpoint evolution and scene configuration become increasingly apparent
over longer horizons.

In contrast, JoyAI-Echo-1.5-Causal exhibits the strongest long-horizon
consistency across these examples. In both natural and structured indoor
environments, it better preserves visual style, scene and subject identity,
and major geometric structures while evolving the viewpoint according to the
prescribed trajectory. These properties remain comparatively stable at later
timestamps instead of progressively degrading with the accumulated
autoregressive history. The qualitative observations are consistent with our
trajectory-following, temporal-degradation, and revisit-consistency metrics,
supporting the improved robustness of our causal model to long-horizon error
accumulation and its stronger action-conditioned world persistence.

\subsubsection{User Study}

We additionally conduct a subjective evaluation on a self-built
World-Model Benchmark (WMB) containing 200 cases.
The benchmark covers game, humanoid-robot, natural outdoor, household
indoor, and driving scenes, with 132 third-person and 68 first-person cases.
We compare JoyAI-Echo-1.5 with LingBot-World-v2~\citep{gao2026lingbot}
and HappyOyster through randomized and anonymized side-by-side comparisons.
Annotators evaluate semantic following, initial-world consistency,
spatiotemporal consistency, motion quality, and visual aesthetics, together
with an overall preference.

\input{tables/tab_gsb_result}

Against LingBot-World-v2, JoyAI-Echo-1.5 receives higher explicit preference
across all five dimensions, with the largest margins in semantic following
(34.0\% vs.\ 15.1\%) and spatiotemporal consistency
(32.6\% vs.\ 14.7\%).
The overall preference is 46.5\% for JoyAI-Echo-1.5 versus 21.6\% for
LingBot-World-v2.

Against HappyOyster, JoyAI-Echo-1.5 shows substantial advantages in
spatiotemporal consistency (57.7\% vs.\ 10.4\%), visual aesthetics
(52.9\% vs.\ 23.8\%), and initial-world consistency
(39.5\% vs.\ 15.4\%).
HappyOyster is preferred more often in semantic following and motion quality,
while JoyAI-Echo-1.5 achieves a substantially higher overall preference
(63.1\% vs.\ 27.1\%).
These results complement the public benchmarks, showing that the gains in
visual quality and spatiotemporal consistency are also reflected in human
preference.

%% file: figures/compare_wbench.tex
\begin{figure}[t!]
    \centering
    \includegraphics[width=\textwidth,height=0.86\textheight,keepaspectratio]{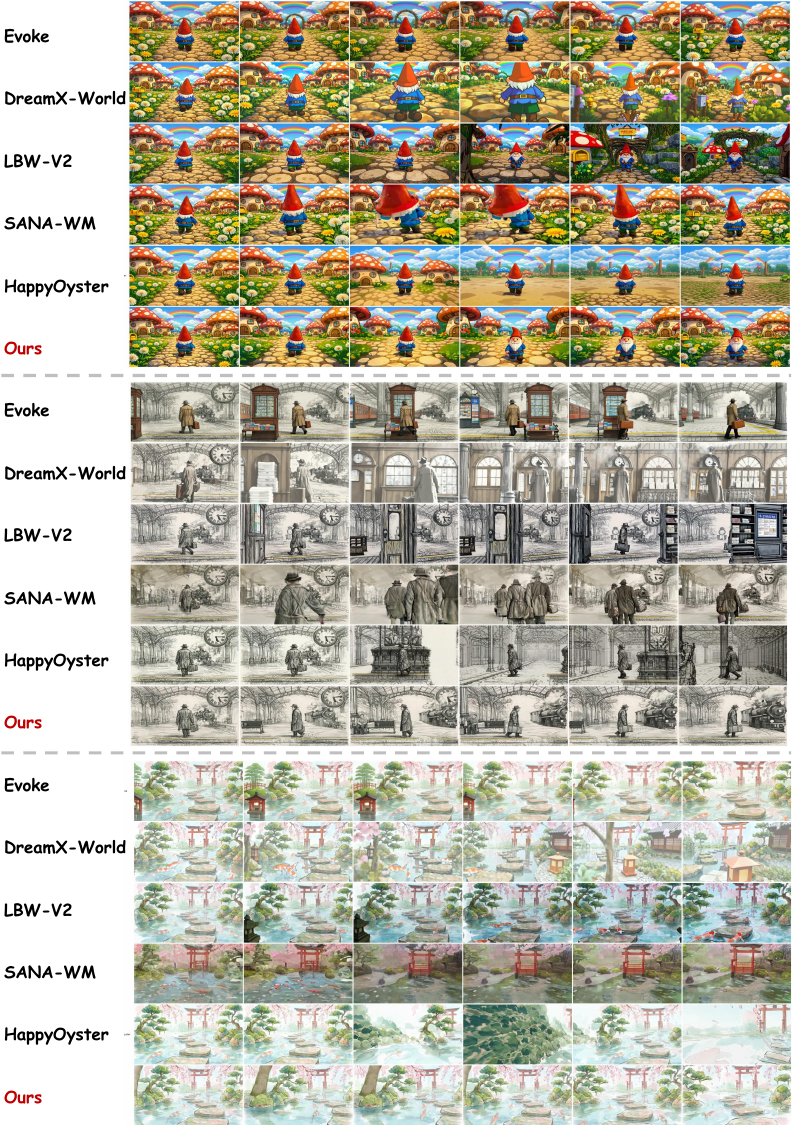}
    \caption{
Qualitative comparison of the undistilled bidirectional multi-step
JoyAI-Echo-1.5 on representative WBench Navigation cases across diverse visual
domains. Given the same initial observation and navigation condition, our model
better preserves scene identity and structure while producing more coherent
viewpoint evolution over successive interaction steps.
}
    \label{fig:wbench-comparison-3}
\end{figure}

%% file: figures/fig_causal_wbench_comparisons.tex
\begin{figure}[t!]
    \centering
    \includegraphics[width=\textwidth,height=0.82\textheight,keepaspectratio]{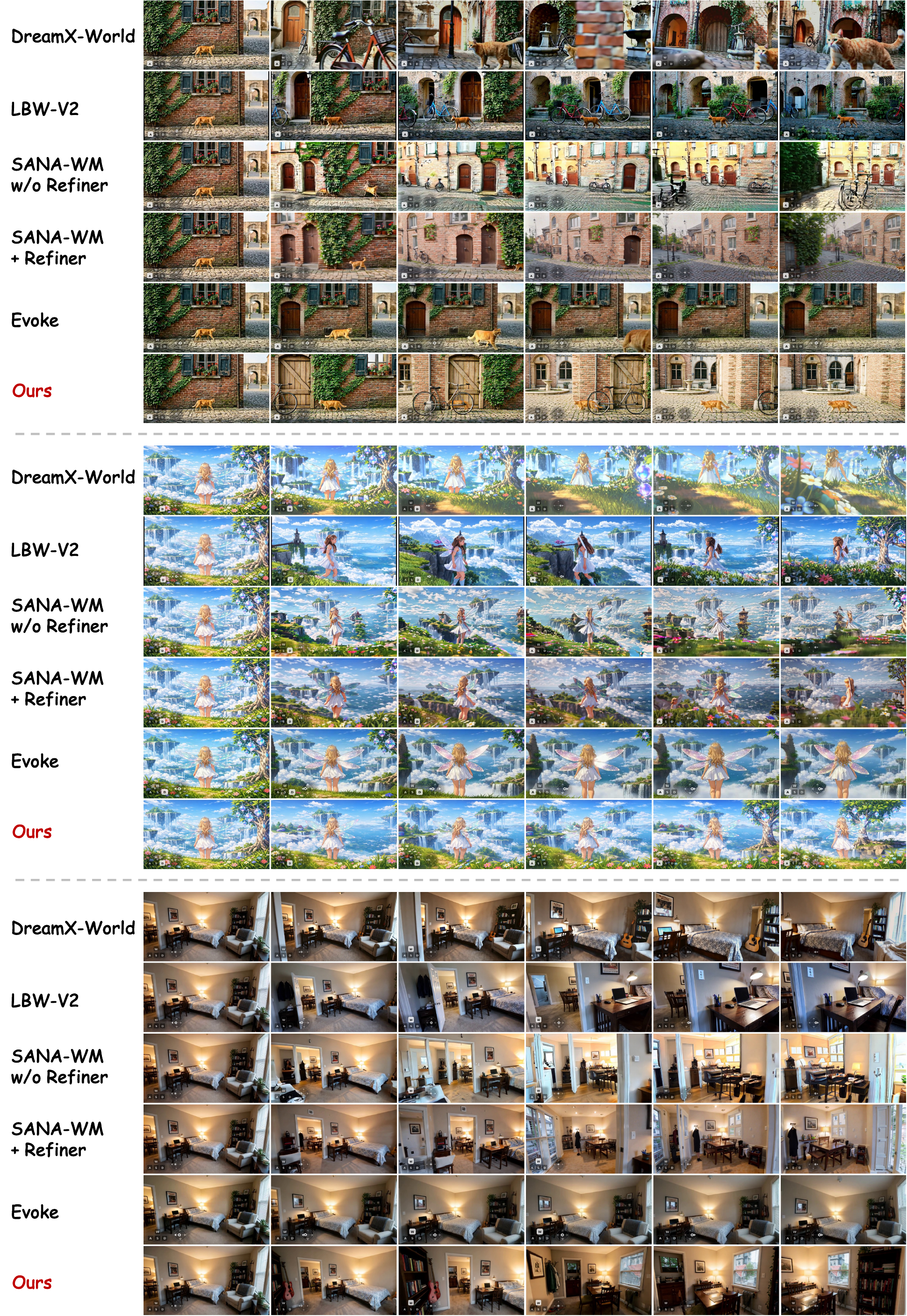}
    \caption{
    Multi-turn comparison of JoyAI-Echo-1.5-Causal against representative
    few-step autoregressive world models on WBench Navigation.
    All models start from the same initial observation and follow the prescribed
    multi-turn action sequence.
    The three cases use
    (1) \texttt{A} $\rightarrow$ \texttt{A} $\rightarrow$
    \texttt{A} $\rightarrow$ \texttt{A},
    (2) \texttt{Left} $\rightarrow$ \texttt{Left} $\rightarrow$
    \texttt{Right} $\rightarrow$ \texttt{Right}, and
    (3) \texttt{Left} $\rightarrow$ \texttt{W} $\rightarrow$
    \texttt{Right}.
    Here, \texttt{Left}, \texttt{Right}, \texttt{Up}, and \texttt{Down}
    denote camera rotations, while \texttt{W}, \texttt{S}, \texttt{A},
    and \texttt{D} denote camera translations.
    JoyAI-Echo-1.5-Causal exhibits stronger action adherence and better
    preserves scene style, subject identity, and world structure across
    successive turns, with less accumulated autoregressive drift.
    }
    \label{fig:wbench-causal-comparison}
\end{figure}

%% file: figures/fig_causal_result_60s.tex
\begin{figure}[t!]
    \centering
    \includegraphics[width=\textwidth,height=0.95\textheight,keepaspectratio]{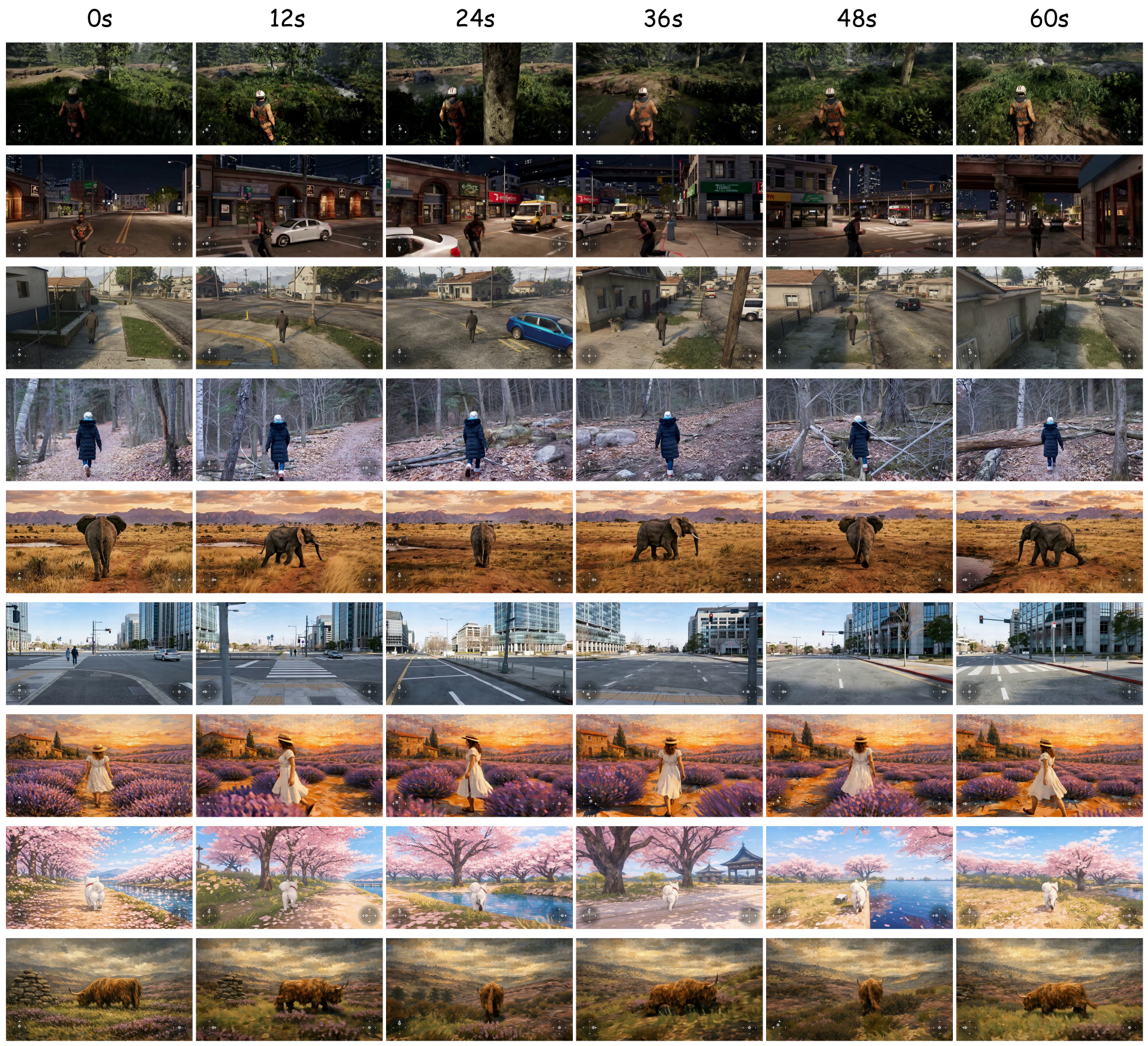}
    \caption{
Causal long-horizon generation of JoyAI-Echo-1.5-Causal on representative cases. Each row shows a 60-second causal rollout sampled at different timestamps. Despite prolonged autoregressive generation, the distilled four-step causal model remains responsive to navigation controls while preserving scene identity, subject appearance, visual style, and major spatial structure, demonstrating robustness to long-horizon error accumulation.
}
    \label{fig:causal-wbench}
\end{figure}

%% file: tables/tab_wbench_navi.tex
\begin{table}[t]
    \centering
    \small
    \caption{
    Comparison on the 158-case WBench Navigation split~\citep{ying2026wbench}.
    JoyAI-Echo-1.5 denotes our undistilled bidirectional multi-step model,
    while JoyAI-Echo-1.5-Causal denotes its distilled four-step causal variant.
    Results are sorted by Average. Best and second-best results are shown in
    bold and underlined, respectively.
    }
    \label{tab:wbench-navigation}

    \resizebox{\linewidth}{!}{%
    \begin{tabular}{lcccccc}
        \toprule
        Model
        & Avg.
        & Quality
        & Setting
        & Interaction
        & Consistency
        & Physical \\
        \midrule

        \textbf{JoyAI-Echo-1.5}
        & \textbf{81.7}
        & 81.5
        & 79.4
        & \underline{87.2}
        & \underline{89.8}
        & 70.6 \\

        \textbf{JoyAI-Echo-1.5-Causal}
        & \underline{81.0}
        & 81.1
        & 88.3
        & \textbf{87.9}
        & 77.5
        & 70.1 \\

        \midrule

        HiDream-O1-World~\citep{ying2026wbench}
        & 80.7
        & 81.9
        & 81.9
        & 79.5
        & 88.0
        & \textbf{72.1} \\

        LingBot-World (fast v2)~\citep{gao2026lingbot}
        & 79.4 & 81.8 & 76.8 & 82.8 & 86.5 & 69.1 \\

        Kling 3.0~\citep{ying2026wbench}
        & 79.0 & 81.4 & 91.0 & 69.4 & 83.7 & 69.3 \\

        LingBot-World (base-camera)~\citep{gao2026lingbot}
        & 78.5 & 78.9 & 72.6 & 80.1 & \textbf{89.9} & 71.2 \\

        Wan 2.7~\citep{wan2025wan}
        & 78.1 & 81.5 & \underline{91.4} & 64.4 & 81.6 & \underline{71.8} \\

        HY-World 1.5 (AR-distill)~\citep{sun2025worldplay}
        & 78.1 & 78.1 & 72.2 & 86.8 & 86.9 & 66.3 \\

        HY-Video 1.5~\citep{wu2025hunyuanvideo15}
        & 77.9 & 77.6 & 85.6 & 71.4 & 87.4 & 67.4 \\

        LingBot-World (fast)~\citep{gao2026lingbot}
        & 77.4 & 79.4 & 77.9 & 79.2 & 84.9 & 65.7 \\

        HappyOyster~\citep{ying2026wbench}
        & 76.8 & 77.3 & 74.2 & 84.9 & 84.3 & 63.5 \\

        Lyra 2.0 (4-step AR)~\citep{shen2026lyra2}
        & 76.4 & 77.1 & 73.2 & 85.6 & 79.3 & 66.7 \\

        Seedance 1.5~\citep{seedance2025seedance}
        & 76.2 & \textbf{82.1} & 82.9 & 66.3 & 81.3 & 68.4 \\

        Cosmos3-Super~\citep{nvidia2026cosmos3}
        & 76.0 & 76.4 & \textbf{91.6} & 58.0 & 85.0 & 69.2 \\

        SANA-WM (4-step AR)~\citep{zhu2026sanawm}
        & 76.0 & 79.3 & 76.1 & 82.2 & 80.7 & 61.9 \\

        DreamX-World (5B AR)~\citep{team2026dreamx}
        & 75.0 & 77.5 & 80.8 & 78.6 & 74.9 & 63.3 \\

        ABot-World~\citep{jiang2026abotworld}
        & 74.7 & 76.8 & 71.4 & 84.0 & 79.5 & 61.7 \\

        Cosmos 2.5~\citep{cosmos_v1}
        & 74.6 & 72.9 & 83.3 & 63.1 & 86.5 & 67.4 \\

        Cosmos3-Nano~\citep{nvidia2026cosmos3}
        & 74.2 & 77.4 & 87.3 & 59.1 & 83.7 & 63.6 \\

        LTX-2.3~\citep{hacohen2026ltx2}
        & 74.2 & 77.1 & 85.2 & 66.4 & 77.2 & 64.9 \\

        Genie 3~\citep{deepmind2025genie3}
        & 73.9 & 75.2 & 72.5 & 73.4 & 82.6 & 65.7 \\

        \bottomrule
    \end{tabular}%
    }
\end{table}

%% file: tables/tab_sana_wm_241.tex
\begin{table}[t]
    \centering
    \scriptsize
    \caption{
    Short-horizon comparison on SANA-WM-Bench using the first 241 frames of
    each trajectory. Each split contains 80 scenes. Best and second-best
    results within each split are shown in bold and underlined, respectively.
    }
    \label{tab:sana-wm-bench-241}

    \begin{tabular}{lcccc}
        \toprule
        & \multicolumn{4}{c}{Trajectory and Visual Quality} \\
        \cmidrule(lr){2-5}

        Method
        & VBench$\uparrow$
        & $R\downarrow$
        & $T\downarrow$
        & CMC$\downarrow$ \\
        \midrule

        \multicolumn{5}{l}{\textit{Simple-Trajectory Split}} \\

        Matrix-Game 3~\citep{wang2026matrixgame3}
        & 79.41
        & 4.474
        & 0.406
        & 0.453 \\

        LingBot-World-v2~\citep{gao2026lingbot}
        & 81.13
        & 3.092
        & 0.282
        & 0.310 \\

        HY-WorldPlay~\citep{sun2025worldplay}
        & 82.05
        & 2.256
        & 0.284
        & 0.298 \\

        DreamX-World~\citep{team2026dreamx}
        & \underline{82.42}
        & 1.839
        & 0.261
        & 0.273 \\

        SANA-WM~\citep{zhu2026sanawm}
        & 81.75
        & \textbf{0.489}
        & \underline{0.194}
        & \underline{0.195} \\

        \midrule
        \textbf{JoyAI-Echo-1.5}
        & \textbf{83.91}
        & \underline{0.523}
        & \textbf{0.160}
        & \textbf{0.162} \\

        \midrule
        \multicolumn{5}{l}{\textit{Hard-Trajectory Split}} \\

        Matrix-Game 3~\citep{wang2026matrixgame3}
        & 79.37
        & 12.819
        & 0.583
        & 0.717 \\

        LingBot-World-v2~\citep{gao2026lingbot}
        & 81.67
        & 6.367
        & 0.357
        & 0.426 \\

        HY-WorldPlay~\citep{sun2025worldplay}
        & 81.24
        & 6.346
        & 0.387
        & 0.437 \\

        DreamX-World~\citep{team2026dreamx}
        & \underline{82.03}
        & 8.804
        & 0.467
        & 0.556 \\

        SANA-WM~\citep{zhu2026sanawm}
        & 81.79
        & \textbf{1.248}
        & \textbf{0.199}
        & \textbf{0.206} \\

        \midrule
        \textbf{JoyAI-Echo-1.5}
        & \textbf{83.96}
        & \underline{1.697}
        & \underline{0.256}
        & \underline{0.266} \\

        \bottomrule
    \end{tabular}
\end{table}

%% file: tables/tab_sana_wm_961.tex
\begin{table}[t]
    \centering
    \scriptsize
    \caption{
    Long-horizon comparison on the full SANA-WM-Bench trajectory under the
    official 961-frame setting (60 seconds at 16 FPS).
    For JoyAI-Echo-1.5, we report the undistilled multi-step model.
    Best and second-best results within each split are shown in bold and
    underlined, respectively.
    }
    \label{tab:sana-wm-bench-961}

    \resizebox{\textwidth}{!}{%
    \begin{tabular}{lcccccccc}
        \toprule
        & \multicolumn{4}{c}{Trajectory and Visual Quality}
        & \multicolumn{3}{c}{Revisit Consistency}
        & \multicolumn{1}{c}{Temporal Quality} \\
        \cmidrule(lr){2-5}
        \cmidrule(lr){6-8}
        \cmidrule(lr){9-9}

        Method
        & VBench$\uparrow$
        & $R\downarrow$
        & $T\downarrow$
        & CMC$\downarrow$
        & PSNR$\uparrow$
        & SSIM$\uparrow$
        & LPIPS$\downarrow$
        & $\Delta$IQ$\downarrow$ \\
        \midrule

        \multicolumn{9}{l}{\textit{Simple-Trajectory Split}} \\

        Infinite-World~\citep{wu2026infiniteworld}
        & 79.18
        & 16.55
        & 1.98
        & 2.08
        & 12.60
        & 0.284
        & 0.595
        & 6.72 \\

        LingBot-World~\citep{gao2026lingbot}
        & \textbf{81.82}
        & 10.47
        & 2.01
        & 2.05
        & \underline{14.59}
        & \textbf{0.366}
        & \textbf{0.394}
        & \underline{0.04} \\

        HY-WorldPlay~\citep{sun2025worldplay}
        & 68.82
        & 17.89
        & 2.36
        & 2.45
        & 12.83
        & 0.321
        & 0.616
        & 23.59 \\

        Matrix-Game 3~\citep{wang2026matrixgame3}
        & 78.53
        & 12.96
        & 1.83
        & 1.92
        & 12.29
        & 0.326
        & 0.553
        & 2.41 \\

        SANA-WM~\citep{zhu2026sanawm}
        & 79.29
        & \underline{7.59}
        & \underline{1.59}
        & \underline{1.63}
        & 14.16
        & 0.333
        & 0.458
        & 3.79 \\

        \midrule
        \textbf{JoyAI-Echo-1.5}
        & \underline{81.36}
        & \textbf{3.22}
        & \textbf{0.918}
        & \textbf{0.927}
        & \textbf{15.10}
        & \underline{0.335}
        & \underline{0.451}
        & \textbf{0.02} \\

        \midrule
        \multicolumn{9}{l}{\textit{Hard-Trajectory Split}} \\

        Infinite-World~\citep{wu2026infiniteworld}
        & 79.51
        & 41.31
        & 2.49
        & 2.84
        & 12.04
        & 0.248
        & 0.617
        & 4.16 \\

        LingBot-World~\citep{gao2026lingbot}
        & \textbf{81.89}
        & 18.99
        & \underline{1.65}
        & 1.81
        & 14.08
        & \underline{0.332}
        & \textbf{0.436}
        & \underline{0.58} \\

        HY-WorldPlay~\citep{sun2025worldplay}
        & 70.46
        & 35.46
        & 2.34
        & 2.64
        & 13.72
        & 0.328
        & 0.654
        & 25.88 \\

        Matrix-Game 3~\citep{wang2026matrixgame3}
        & 78.79
        & 18.79
        & 1.67
        & 1.82
        & 12.17
        & 0.317
        & 0.556
        & \textbf{0.32} \\

        SANA-WM~\citep{zhu2026sanawm}
        & 79.60
        & \textbf{10.02}
        & 1.66
        & \underline{1.72}
        & \underline{14.10}
        & 0.327
        & \underline{0.469}
        & 3.09 \\

        \midrule
        \textbf{JoyAI-Echo-1.5}
        & \underline{81.72}
        & \underline{12.05}
        & \textbf{1.265}
        & \textbf{1.358}
        & \textbf{14.58}
        & \textbf{0.333}
        & 0.482
        & 0.93 \\

        \bottomrule
    \end{tabular}%
    }
\end{table}

%% file: tables/tab_sana_wm_961_causal_few_steps.tex
\begin{table}[t]
    \centering
    \scriptsize
    \caption{
    Long-horizon comparison of distilled causal world models on SANA-WM-Bench
    under the 961-frame setting (60 seconds at 16 FPS).
    JoyAI-Echo-1.5-Causal denotes ours 4-step causal model.
    Best and second-best results within each split are shown in bold and
    underlined, respectively.
    }
    \label{tab:sana-wm-bench-causal-961}

    \resizebox{\textwidth}{!}{%
    \begin{tabular}{lcccccccc}
        \toprule
        & \multicolumn{4}{c}{Trajectory and Visual Quality}
        & \multicolumn{3}{c}{Revisit Consistency}
        & \multicolumn{1}{c}{Temporal Quality} \\
        \cmidrule(lr){2-5}
        \cmidrule(lr){6-8}
        \cmidrule(lr){9-9}

        Method
        & VBench$\uparrow$
        & $R\downarrow$
        & $T\downarrow$
        & CMC$\downarrow$
        & PSNR$\uparrow$
        & SSIM$\uparrow$
        & LPIPS$\downarrow$
        & $\Delta$IQ$\downarrow$ \\
        \midrule

        \multicolumn{9}{l}{\textit{Simple-Trajectory Split}} \\

       Evoke~\cite{yin2026alaya}
        & \underline{80.11}
        & \underline{9.859}
        & \underline{2.159}
        & \underline{2.190}
        & \underline{13.50}
        & \underline{0.2774}
        & \underline{0.5010}
        & \underline{-0.49} \\

        LingBot-World-v2~\citep{gao2026lingbot}
        & 79.20
        & 22.516
        & 2.391
        & 2.541
        & 11.90
        & 0.1715
        & 0.5906
        & 2.59 \\

        SANA-WM + Causal Refiner~\citep{zhu2026sanawm}
        & 79.55
        & 17.875
        & 2.222
        & 2.338
        & \textbf{14.41}
        & 0.2660
        & 0.5748
        & \textbf{-0.75} \\

        SANA-WM w/o Refiner~\citep{zhu2026sanawm}
        & 78.21
        & 15.359
        & 2.288
        & 2.393
        & 9.19
        & 0.1661
        & 0.6321
        & -0.12 \\

        \midrule
        \textbf{JoyAI-Echo-1.5-Causal}
        & \textbf{80.13}
        & \textbf{5.009}
        & \textbf{1.592}
        & \textbf{1.611}
        & \textbf{14.41}
        & \textbf{0.2998}
        & \textbf{0.4802}
        & 2.09 \\

        \midrule
        \multicolumn{9}{l}{\textit{Hard-Trajectory Split}} \\

        Evoke~\cite{yin2026alaya}
        & \underline{80.75}
        & 20.097
        & \underline{1.924}
        & \underline{2.060}
        & 13.06
        & 0.2592
        & \underline{0.5292}
        & \underline{0.30} \\

        LingBot-World-v2~\citep{gao2026lingbot}
        & 78.74
        & 34.416
        & 2.125
        & 2.413
        & 12.34
        & 0.2183
        & 0.5732
        & 3.84 \\

        SANA-WM + Causal Refiner~\citep{zhu2026sanawm}
        & 79.60
        & 23.794
        & 1.932
        & 2.124
        & \underline{13.89}
        & \underline{0.2605}
        & 0.5696
        & \textbf{-0.24} \\

        SANA-WM w/o Refiner~\citep{zhu2026sanawm}
        & 78.27
        & \underline{18.717}
        & 2.006
        & 2.142
        & 9.26
        & 0.1729
        & 0.6077
        & 2.06 \\

        \midrule
        \textbf{JoyAI-Echo-1.5-Causal}
        & \textbf{81.06}
        & \textbf{14.221}
        & \textbf{1.735}
        & \textbf{1.829}
        & \textbf{14.04}
        & \textbf{0.2910}
        & \textbf{0.4893}
        & 1.30 \\

        \bottomrule
    \end{tabular}%
    }
\end{table}

%% file: figures/fig_sana_iq_summary.tex
\begin{figure}[t!]
    \centering
    \includegraphics[width=\textwidth]{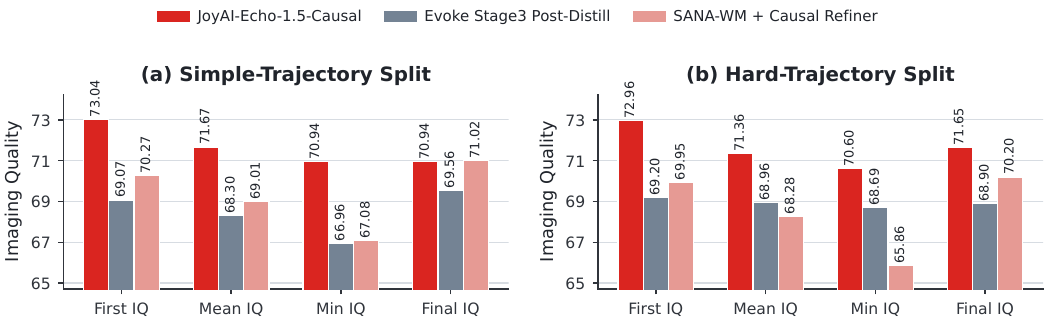}
    \caption{Imaging quality comparison on the Simple- and Hard-Trajectory splits. We report the first-window, mean, minimum, and final-window IQ. Our method achieves consistently higher mean and minimum IQ than Evoke and SANA-WM with Causal Refiner. Although the competing methods exhibit smaller IQ drops, their overall IQ remains lower, indicating that a small drop can also result from consistently low imaging quality rather than stronger long-horizon quality preservation.}
    \label{fig:iq-comparison}
\end{figure}

%% file: figures/fig_sana_causal_results_comparisons.tex
\begin{figure}[t!]
    \centering
    \includegraphics[width=0.96\textwidth,height=0.84\textheight,keepaspectratio]{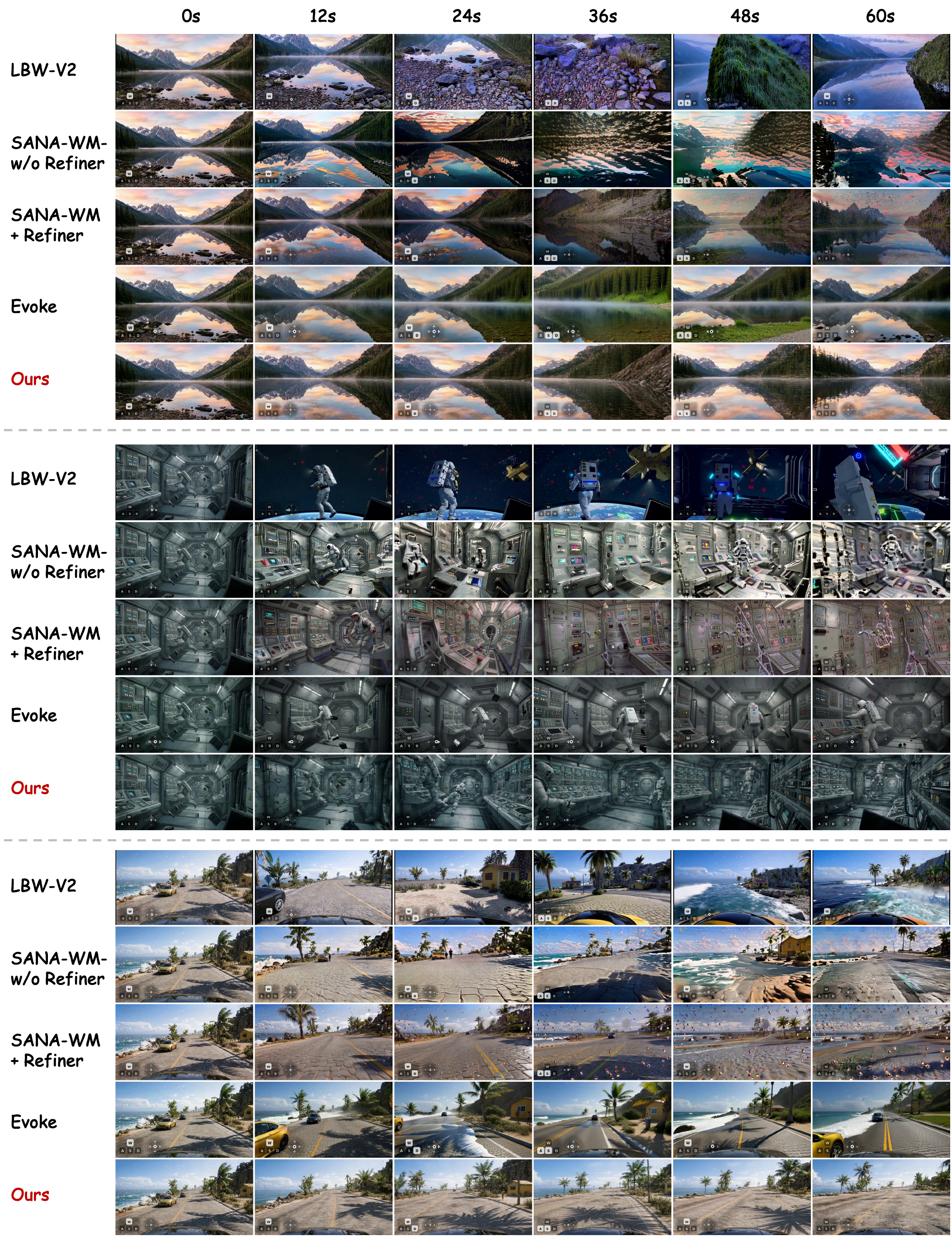}
    \caption{
    Qualitative comparison of 961-frame (60-second at 16 FPS) causal rollouts
    on representative SANA-WM-Bench trajectories. We compare
    JoyAI-Echo-1.5-Causal with LingBot-World-v2 (LBW-V2),
    SANA-WM with and without Causal Refiner, and Evoke.
    Frames are sampled every 12 seconds from the same initial observation and
    camera-trajectory condition. Across the extended rollout,
    JoyAI-Echo-1.5-Causal exhibits less accumulated autoregressive drift,
    more consistent trajectory following, and stronger preservation of scene
    identity, visual style, and spatial structure.
    }
    \label{fig:sana-causal-qualitative}
\end{figure}

%% file: tables/tab_gsb_result.tex
\begin{table}[t]
\centering
\caption{
Human evaluation based on Good--Similar--Bad (GSB) pairwise comparisons
between JoyAI-Echo-1.5 (world model variant) and Happy Oyster for world modeling.
Numbers denote the percentage of user preferences.
}
\label{tab:wm_gsb_happyoyster}
\begin{tabular}{lccc}
\toprule
Aspect & JoyAI-Echo-1.5 & Similar & Happy Oyster \\
\midrule
Semantic following
    & 13.63\% & 56.94\% & \textbf{29.44\%} \\
Initial-world consistency
    & \textbf{39.50\%} & 45.06\% & 15.44\% \\
Motion
    & 18.94\% & 56.51\% & \textbf{24.56\%} \\
Spatiotemporal consistency
    & \textbf{57.69\%} & 31.88\% & 10.44\% \\
Visual aesthetics
    & \textbf{52.88\%} & 23.32\% & 23.81\% \\
Overall
    & \textbf{63.13\%} & 9.82\% & 27.06\% \\
\bottomrule
\end{tabular}
\end{table}

\begin{table}[t]
\centering
\caption{
Human evaluation based on Good--Similar--Bad (GSB) pairwise comparisons
between JoyAI-Echo-1.5 (world model variant) and LingBot-World-v2 for world modeling.
Numbers denote the percentage of user preferences.
}
\label{tab:wm_gsb_lingbot}
\begin{tabular}{lccc}
\toprule
Aspect & JoyAI-Echo-1.5 & Similar & LingBot-World-v2 \\
\midrule
Semantic following
    & \textbf{34.00\%} & 50.86\% & 15.14\% \\
Initial-world consistency
    & \textbf{14.64\%} & 77.65\% & 7.71\% \\
Motion
    & \textbf{25.29\%} & 54.71\% & 20.00\% \\
Spatiotemporal consistency
    & \textbf{32.64\%} & 52.65\% & 14.71\% \\
Visual aesthetics
    & \textbf{32.50\%} & 47.57\% & 19.93\% \\
Overall
    & \textbf{46.50\%} & 31.92\% & 21.57\% \\
\bottomrule
\end{tabular}
\end{table}

%% file: conclusion.tex
\section{Conclusion}

We presented a framework for extending audio-visual generation beyond isolated clips toward persistent stories and interactive worlds. The long-video system uses composable cross-shot audio-visual memory to preserve character appearance, speaker identity, and narrative continuity under flexible text, image, and memory conditioning. The world-model system translates heterogeneous navigation inputs into calibrated metric 6-DoF trajectories, enabling precise and controller-agnostic interaction across diverse environments and viewpoints. Together with few-step distillation, causal audio-visual teacher forcing, and short- and long-horizon Self-Gradient Forcing, these designs allow the models to generate synchronized audio and video efficiently while remaining stable under self-generated histories. Experiments demonstrate improved cross-shot consistency, instruction following, speech fidelity, trajectory control, and long-horizon world persistence, with leading results on both long-form generation evaluations and public world-model benchmarks. These findings suggest that memory, geometric control, and rollout-aware training are complementary foundations for video systems that can remember past events, respond to user intent, and continue evolving over time. Accumulated rotational drift on difficult long-horizon trajectories remains a key limitation, motivating future work on stronger geometric consistency, persistent world-state representations, and more reliable open-ended interaction.